%% file: camera-ready.tex
\documentclass[sigconf]{acmart}
\renewcommand\footnotetextcopyrightpermission[1]{} 
\usepackage{multirow}

\usepackage{amssymb}
\usepackage{bbding}
\usepackage{colortbl}
\usepackage{xcolor}
\usepackage{stfloats}

\AtBeginDocument{%
  \providecommand\BibTeX{{%
    \normalfont B\kern-0.5em{\scshape i\kern-0.25em b}\kern-0.8em\TeX}}}

\setcopyright{acmlicensed}
\copyrightyear{2018}
\acmYear{2018}
\acmDOI{XXXXXXX.XXXXXXX}

\acmConference[MM'24]{}
\acmISBN{}

\begin{document}

\title{AutoTVG: A New Vision-language Pre-training Paradigm for Temporal Video Grounding}

\author{Xing Zhang$^{1}$ \quad Jiaxu Gu$^{2}$ \quad Haoyu Zhao$^1$ \quad Shicong Wang$^1$ \quad Hang Xu$^{2}$ \quad Renjing Pei$^{2}$ \quad \\
Songcen Xu$^{2}$ \quad Zuxuan Wu$^{1}$ \quad Yu-Gang Jiang$^{1}$ \\
    $^1$Fudan University, China\\ 
    $^2$Huawei Noah’s Ark Lab, China
   }

\renewcommand{\shortauthors}{author name and author name, et al.}

\begin{abstract}

\input{abstract}
\end{abstract}



\maketitle

\input{introduction}

\input{related_work}

\input{method}

\input{experiments}

\input{conclusion}

\bibliographystyle{ACM-Reference-Format}
\bibliography{sample-base}

\clearpage

\appendix

\section*{\centering Appendix}

We provide more details about AutoTVG in the appendix including

\begin{itemize}

    \item Analysis of pre-training dataset in Section~\ref{sec:ht100m}.
    
    \item Limitations of CMG (Captioned Moment Generation) in Section~\ref{sec:limitation}.

\end{itemize}

\input{ht100m.tex}

\input{limitation.tex}










\end{document}

%% file: abstract.tex
Temporal Video Grounding (TVG) aims to localize a moment from an untrimmed video given the language description. Since the annotation of TVG is labor-intensive, TVG under limited supervision has accepted attention in recent years. The great success of vision-language pre-training guides TVG to follow the traditional ``pre-training + fine-tuning'' paradigm, however, the pre-training process would suffer from a lack of temporal modeling and fine-grained alignment due to the difference of data nature between pre-train and test. Besides, the large gap between pretext and downstream tasks makes zero-shot testing impossible for the pre-trained model.
To avoid the drawbacks of the traditional paradigm, we propose AutoTVG, a new vision-language pre-training paradigm for TVG that enables the model to learn semantic alignment and boundary regression from automatically annotated untrimmed videos. 
To be specific, AutoTVG consists of a novel Captioned Moment Generation (CMG) module to generate captioned moments from untrimmed videos, and TVGNet with a regression head to predict localization results. Experimental results on Charades-STA and ActivityNet Captions show that, regarding zero-shot temporal video grounding, AutoTVG achieves highly competitive performance with in-distribution methods under out-of-distribution testing, and is superior to existing pre-training frameworks with much less training data.

%% file: introduction.tex
\section{Introduction}


\begin{figure}[t]
    \centering
    \includegraphics[width=0.98\linewidth]{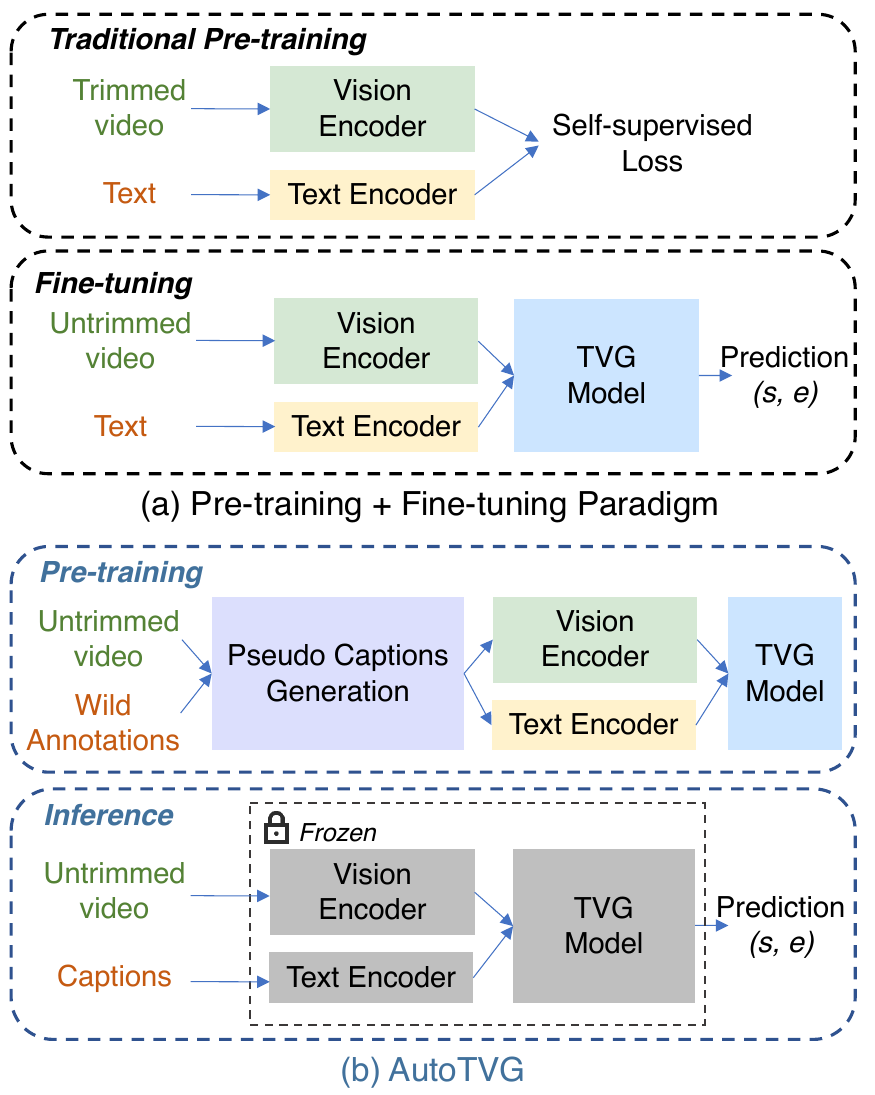}
    \caption{Comparisons of the traditional ``pre-training + fine-tuning'' paradigm and the proposed AutoTVG. The traditional method follows a two-step strategy which pre-trains vision and text encoders with self-supervised loss and then fine-tunes a TVG model, while the proposed AutoTVG pre-trains encoders and a TVG model in a single step with untrimmed videos, so that can perform zero-shot testing.}
    \label{fig:intro}
\end{figure}

Temporal Video Grounding (TVG) aims to localize a temporal segment from an untrimmed video, which is most related to the given query~\cite{yuan2019find,mun2020local,zhang2020learning,li2022compositional,ding2021support,wang2021structured,zhang2019man, yuan2021closer, wang2023mixup, chen2023curriculum}. However, the labor-intensive annotation process of captions and timestamps makes TVG not applicable to real-world scenarios. To this end, researchers devote themselves to developing TVG systems with less supervision, such as weakly-supervised TVG~\cite{duan2018weakly,lin2020weakly,song2020weakly,gao2020weakly,huang2021cross,ma2020vlanet} which removes the timestamps of captions and zero-shot TVG~\cite{nam2021zero,wang2022prompt} which only utilizes videos for training.

To meet the urgent need of TVG from the industry, reducing annotation costs is by no means a worthy challenge to address. In the past few years, the pre-training technique has raised much attention from both academia and industry, and is regarded as a potential way of saving manual labeling. For instance, CLIP~\cite{radford2021learning} utilizes 4 Million image-text paired data to pre-train a dual-encoder model, it reveals that self-supervised from paired multi-modal data can improve the generalization ability of model representations, so that it can perform well on zero-shot classification and retrieval. 

Inspired by the great success of pre-training works, video-language pre-training for video localization has attracted attention from researchers as well. LocVTP~\cite{cao2022locvtp} proposes the first pre-training framework to solve TVG and temporal action localization, however, the traditional ``pre-training + fine-tuning'' paradigm (see Figure~\ref{fig:intro}) has some obvious drawbacks to TVG. 
Firstly, the typical pre-training video dataset consists of trimmed short-range videos while the videos from TVG datasets are untrimmed long-range videos, e.g. WebVid-2M \textit{vs.} ActivityNet, which would result in two major limitations in the pre-training process: 1) \textbf{\textit{poor temporal modeling}}: the lack of motion dynamics in trimmed videos are unlikely to make the pre-trained model obtain temporal discriminative representations; 2) \textit{\textbf{low fine-grained alignment}}: the caption for a short video may be an overall summary which is related to each frame, so that makes it hard to construct positive and negative samples for mining fine-grained alignment with frame-by-word comparison.
Secondly, the gap between self-supervised pretext task of pre-training and downstream localization task is non-negligible, which makes it hard to achieve zero-shot testing for TVG as easily as classification and retrieval tasks.

In order to take full advantage of pre-training models' merits, we propose a new paradigm of pre-training for TVG. As Figure~\ref{fig:intro} shows, we utilize untrimmed videos with automatically generated annotations (i.e., subtitles and timestamps) as raw input, then a novel Captioned Moment Generation (CMG) module is applied to generate candidate moments from videos and captions from subtitles, in which a CLIP model is used to retrieve nouns and verbs that best match video moments. 
Through experiments, we find that the training data with raw subtitles and corresponding timestamps are inferior for model convergence, while the generated captioned moments are capable of pre-training with CMG to refine the visual-text alignment and denoise the words in subtitles. 
Compared to previous zero-shot methods, the diversity of nouns and verbs from subtitles brings the advantage of generalization ability to TVG model, which is essential for out-of-distribution testing.
In total, the new paradigm addresses the aforementioned problems from two aspects. On the one hand, we narrow the gap between pre-training and downstream datasets by introducing untrimmed videos for training, which enables the model to capture temporal differences among frames. On the other hand, we bridge the pre-training and fine-tuning processes by explicitly pre-training a TVG model with pseudo labels, therefore, the pre-trained model is capable of zero-shot testing on TVG datasets.

To sum up, our contributions are as follows:
\begin{itemize}

\item We propose an effective pre-training framework to tackle temporal video grounding, with generated captioned moments from untrimmed videos, the pre-trained model can perform zero-shot testing on downstream TVG datasets.

\item We provide an effective Captioned Moment Generation module to generate captioned video moments from the vision-language pre-training model, extensive experiments illustrate the diversity of nouns and verbs of training data can provide strong generalization ability to the TVG model, which can inspire further research on the generalization of TVG models.

\item Despite the training data being out-of-distribution from testing data, the proposed method achieves highly comparable results with existing in-distribution zero-shot methods or even surpasses some weakly-supervised methods on two standard TVG datasets including Charades-STA and ActivityNet Captions. Compared to the pre-training method for temporal video grounding, our framework achieves better performance with much less training data.
\end{itemize}


%% file: related_work.tex
\section{Related Work}
\label{sec:related_work}

\subsection{Vision-Language Pre-training (VLP).}
With the success of pre-training methods in computer vision and natural language processing, the ``pretraining-then-finetuning'' paradigm has been applied to various VLP tasks. 
Vision-language models obtain strong representation ability through pre-training on large-scale image-text pairs with self-supervision~\cite{radford2021learning,zhou2022learning,zhou2022conditional,rao2022denseclip,li2022grounded,li2022blip}. 
Such paradigm has been applied to various applications, including zero-shot image classification, cross-modal retrieval, etc. As one of the most representative VLP method is CLIP~\cite{radford2021learning} which pre-trains a dual-encoder model through self-supervised contrastive learning on 4M image-text paired dataset. 

With the success of pre-trained vision-language models on image-text matching tasks such as classification and retrieval, researchers then adapt pre-trained models to dense prediction tasks. 
DenseCLIP~\cite{rao2022denseclip} converts the original image-text matching scheme in CLIP to pixel-text matching. 
GLIP~\cite{li2022grounded} aligns image regions with object labels in a way of grounding to enhance detection as well as grounding.
For video retrieval task ~\cite{lei2021understanding,akbari2021vatt,sun2019videobert,li2020hero,xu2021vlm}, VideoBERT~\cite{sun2019videobert} is the first work of video-language pre-training, which aims to mine high-level semantic information in videos. Besides, a series of works have been proposed to address QA tasks as well~\cite{tang2021decembert,lei2021less,li2020hero,xu2021vlm,zhu2020actbert}.
HERO~\cite{li2020hero} introduces a hierarchical Transformer to model the temporal information of videos and designs two pre-training tasks. DeCEMBERT~\cite{tang2021decembert} alleviates the problem of alignment bias in captions and videos by designing a constrained attention loss to select the most matching caption.
Although there are many works that apply vision-language pre-training to image dense prediction tasks, e.g., object detection, segmentation, and image grounding, limited efforts have been made to video localization.

\vspace{-3mm}
\subsection{Temporal Video Grounding (TVG).}
TVG is described as finding the specific video clip that is best described by a query. It was proposed in 2017 by~\cite{anne2017localizing,gao2017tall} and has attracted increasing attention from academia.
Existing TVG methods can be divided into three categories: fully-supervised, weakly-supervised and unsupervised. Most existing methods focus on fully-supervised settings~\cite{yuan2019find,mun2020local,zhang2020learning,li2022compositional,ding2021support,wang2021structured,zhang2019man} to model the semantic relationship between text and video. 
However, dense annotations of timestamps and captions for supervised learning are expensive. To avoid the intensive labor involved in regional annotations, weakly-supervised TVG has been introduced. Existing weakly-supervised TVG works mainly fall into two categories, one is reconstruction-based methods~\cite{duan2018weakly,lin2020weakly,song2020weakly,yang2021local} which generate timestamps and captions of segments through cyclic training,
the other is multi-instance learning-based methods~\cite{gao2020weakly,huang2021cross,ma2020vlanet,mithun2019weakly,zhang2020counterfactual,tan2021logan} which minimizes score between negative samples and maximizes score between positive samples to learn fine-grained alignment between video and query.

To further eliminate the dependence on data annotations, PSVL~\cite{nam2021zero} introduces zero-shot TVG which only takes videos for training, with timestamps and captions removed. PSVL first generates pseudo labels for videos which are used to train a TVG model afterward, then in-distribution zero-shot test is performed for evaluation. A follow-up work is PZVMR~\cite{wang2022prompt}, which replaces the process of verbs in PSVL with prompt learning.

However, previous methods generate nouns with off-the-shelf object detection models which are limited in word diversity, our AutoTVG obtains nouns and verbs from ASR or user-uploaded captions which contain more action or object instances, so the pre-trained model can generalize to out-of-distribution test data.

\vspace{-3mm}
\subsection{Video Pre-training for Temporal Localization (VPTL).}
VPTL aims to apply the ``pre-training + fine-tuning'' paradigm to temporal localization task. Existing VPTL works mainly focus on the pure vision domain~\cite{alwassel2021tsp,xu2021boundary,xu2021low,zhang2022unsupervised}. 
Recently, the first vision-language pre-training framework for temporal localization is introduced~\cite{cao2022locvtp}, which joint trains video and text encoders with pretext tasks including video-sentence contrastive learning, clip-word contrastive learning and temporal reasoning learning. However, it has two major limitations. Firstly, the pre-training process is based on trimmed video dataset WebVid-2M, the lack of motion dynamics in trimmed videos may result in poor temporal modeling for encoders,  also the caption may be an overall summary related to each frame so fine-grained alignment is hard to achieve; Secondly, the huge gap between pretext task and downstream task makes it impossible for the pre-trained model to perform zero-shot testing. Another progress is UniVTG~\cite{lin2023univtg} which unifies diverse temporal video grounding labels and tasks to achieve zero-shot temporal video grounding, however, the performance is still limited under large-scale pre-training.

Considering the limitations of ``pre-training + fine-tuning'' paradigm for TVG, we propose a new paradigm AutoTVG which narrows the gap between dataset and task for pre-training and testing, so that the pre-trained model can be directly used for zero-shot test.

%% file: method.tex
\section{Approach}
\label{sec:method}

\subsection{Problem Formulation}
Temporal Video Grounding (TVG) aims at locating a time period from an untrimmed video, during which the video content is semantically relevant to a given text query. 
Due to the intensive labor of annotation, weakly-supervised and zero-shot TVG have received a lot of attention in the past few years. Weakly-supervised TVG only uses videos and queries as input, while only videos are available for zero-shot TVG. 
In this paper, we mainly conduct experiments under zero-shot setup.

\subsection{Method Overview}

\begin{figure}[t!]
    \centering
    \includegraphics[width=0.99\linewidth]{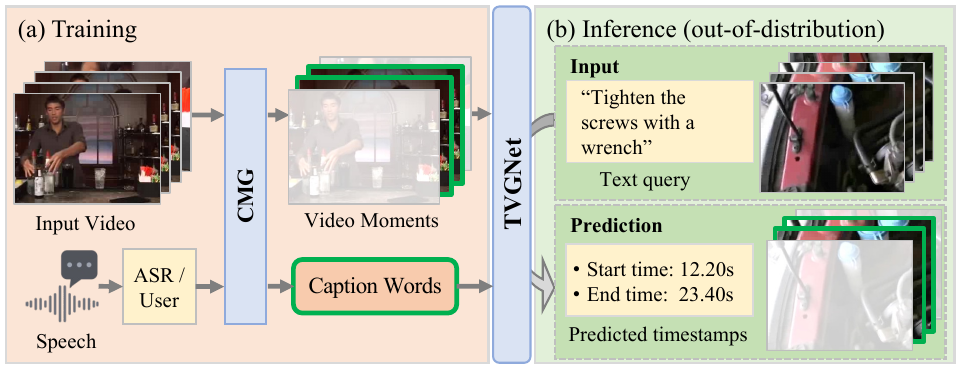}
    \caption{An overview of our proposed method AutoTVG which consists of two main modules CMG and TVGNet. CMG module is for generating captioned moments from untrimmed videos by exploiting the speech in videos, the generated captioned moments are utilized for pre-training TVGNet. }
    \label{fig:method_pipeline}
\end{figure}

Figure~\ref{fig:method_pipeline} shows the overview of our proposed method AutoTVG. We take untrimmed and unannotated video dataset for model pre-training, which consists of videos with subtitles automatically generated by software like Automatic Speech Recognition (ASR) engines or users. The basic idea of our method is to generate captioned moments from a dataset without human-labeled annotations and then pre-train a general TVG model for downstream evaluation. Specifically, our framework contains two main modules; 1) Captioned Moment Generation (CMG)  for generating captioned video moments from untrimmed videos; 2) Temporal Video Grounding Network (TVGNet) with a regression head to learn boundary prediction from the generated caption moments. After pre-training, TVGNet can be directly adapted to downstream TVG datasets in a zero-shot manner.

\subsection{Captioned Moment Generation (CMG)}

\begin{figure*}[t]
    \centering
    \includegraphics[width=0.96\linewidth]{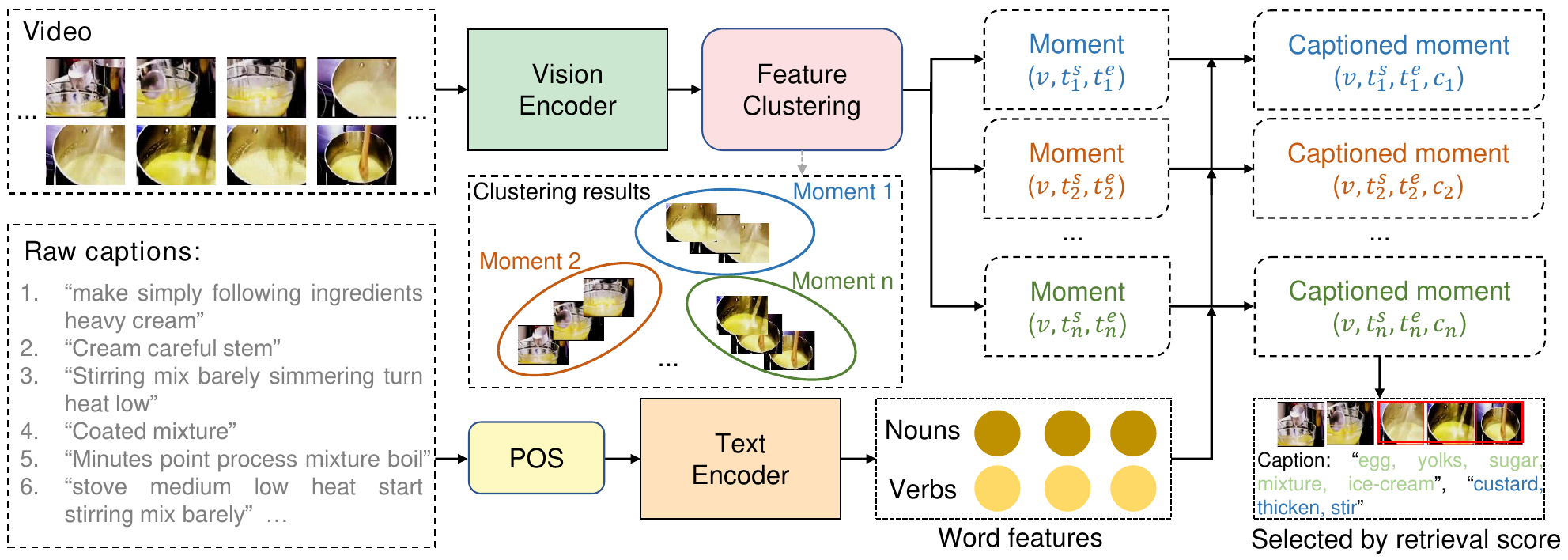}
    \caption{The pipeline of CMG. Video moments are generated by clustering video frame features and their corresponding captions are decided by cross-modal alignment. To reduce the impact of noise from raw captions, only nouns and verbs are picked out to caption the resulting moments.}
    \label{fig:cmg}
\end{figure*}

Figure~\ref{fig:cmg} shows the structure of Captioned Moment Generation (CMG) module. For an untrimmed video dataset, videos are denoted as $\mathcal{V}$ and their corresponding automatically generated subtitles are denoted as $\mathcal{C}$. The size of the dataset is denoted as $N$ so we have $N$ pairs of video and text captions: $(V_i, C_i), i \in \left\{1, 2, \ldots, N\right\}$. For each pair, the video consists of $P$ frames $\left\{f_1, f_2, \ldots, f_p, \ldots, f_P\right\}$ and the caption is decomposed into $Q$ words $\left\{w_1, w_2, \ldots, w_q, \ldots, w_Q\right\}$ by Part-of-Speech Tagging~\cite{spacy}.

Each pair of video and text caption $\left\{V_i, C_i\right\}$ is considered as an input of the CMG module. We exploit CLIP~\cite{radford2021learning} as the image encoder for frame encoding, which is a powerful dual-encoder multi-modal model trained on large-scale image-text pairs. Then we conduct Video Moment Generation on frame features to obtain video moments. We also exploit CLIP text encoder to encode word features and then the Moment Caption Selection is applied based on frame features and word features, which will be introduced in detail as below.

\noindent
\textbf{Video Moment Generation.} 
Given the extracted video frame features $\left\{F_1, F_2,..., F_p,..., F_P\right\}$ and frame indices $\left\{1, 2,..., p,..., P\right\}$, we use K-means clustering algorithm for feature clustering to obtain candidate moments $\left\{(v, t_1^s, t_1^e), (v, t_2^s, t_2^e),...,(v, t_n^s, t_n^e)\right\}$ by taking the concatenated vector of frame features and frame indices as input.
The frame indices are crucial for obtaining continuous events as a candidate moment, since it encourages adjacent frames to be clustered in one moment, more details could be found in the supplementary.

After obtaining candidate moments, several candidate selection methods including random, longest and distinct are applied to choose one moment as the final output. In practice we find random performs best, see experiments for more details.



\noindent
\textbf{Moment Caption Selection.} Since the automatically generated subtitles are too noisy to TVG task, we try to select a few representative words to construct a ''clean'' caption for a video moment. After Part-of-Speech Tagging, the subtitles are divided into a set of $Q$ words $\left\{w_1, w_2, \ldots, w_q, \ldots, w_Q\right\}$ from which we obtain $I$ candidate nouns $\left\{noun_1, noun_2, \ldots, noun_i, \ldots, noun_I\right\}$ and $J$ candidate verbs $\left\{verb_1, verb_2, \ldots, verb_j, \ldots, verb_J\right\}$.

Then we exploit CLIP text encoder to obtain noun and verb features, the intuition is that CLIP provides the function of image-text alignment since it is pre-trained on large-scale image-text pairs, so that it could help select words that can match a video moment.
Inspired by CLIPScore~\cite{hessel2021clipscore}, a straight forward reference-free evaluation metric for image captioning, we evaluate each word by its similarity with video moment, then take top $N_1$ nouns and top $N_2$ verbs from candidate nouns and verbs. As a result, the selected nouns and verbs are concatenated to construct the final caption $c$. 

After video moment generation and moment caption selection, we obtain captioned moments $\left\{(v, t_1^s, t_1^e, c_1), (v, t_2^s, t_2^e, c_2),...,(v, t_n^s, t_n^e, \right. \\ \left. c_n)\right\}$ which are used to pre-train a temporal video grounding network in the following.

\subsection{Temporal Video Grounding Network}

\begin{figure}[h]
    \centering
    \includegraphics[width=0.96\linewidth]{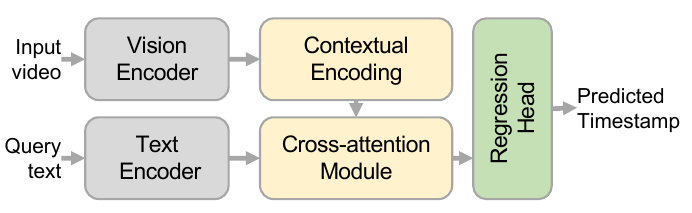}
    \caption{The structure of TVGNet. Two encoders are used for extracting features from input video and query text respectively, a contextual encoding module is for modeling global context-aware features, and cross-modal features are integrated through a cross-attention module. Finally, a timestamp is predicted by a regression head.}
    \label{fig:tvgnet}
\end{figure}

The temporal video grounding network (TVGNet) is to fuse videos and captions as well as predict time intervals from videos that match their captions.
A typical TVGNet structure is illustrated in Figure~\ref{fig:tvgnet}.
We take the basic structure as~\cite{nam2021zero} that contains a contextual encoding module to model global context-aware video features; a cross-attention model containing word-guided attention, video-guided attention, and multi-modal cross-attention blocks to fuse video and text features; a regression head with multi-layer perceptron layers to predict boundaries of video moments. As for word embedding, different from previous works, we abandon GloVe embedding for generated moment captions since it relies on a vocabulary dictionary built from the word frequency of training data. We argue that the embedding method built upon training word frequency is not suitable for zero-shot scenario, since the vocabulary is different between pre-train data and downstream data. To this end, we take CLIP text encoder to obtain text features and remove text contextual encoding.

After pre-training TVGNet, the pre-trained model can directly adapt to downstream grounding datasets. The following experimental results show the effectiveness of the pre-training process.

\noindent
\textbf{Loss function.} Following~\cite{nam2021zero}, the proposed method consists of two loss functions; 1) temporal boundary regression loss $\mathcal L_{reg}$ and 2) temporal attention guided loss $\mathcal L_{guide}$:
\begin{equation}
\begin{aligned}
\mathcal L_{total} = \mathcal L_{reg} + \lambda\mathcal L_{guide}
\end{aligned}
\end{equation}

Where $\lambda$ is a balancing parameter, Considering the coarse alignment between caption words and video content, less sensitivity to outliers is preferred so the Huber loss function is used for $\mathcal L_{reg}$.

%% file: experiments.tex
\section{Experiments}
\label{sec:experiments}

\subsection{Experimental settings}

\noindent\textbf{Pre-training dataset.} We take a subset of HowTo100M by random sampling 30K videos from 1.2M videos. HowTo100M~\cite{miech2019howto100m} is a large-scale instructional video dataset that contains 136M video clips from 1.2M YouTube videos and 23K diverse activities from multiple domains, each video has rich subtitles which are commonly automatically generated via ASR or users.
Although subtitles provide both text and timestamps for a short video clip, however, they are rather noisy and redundant which is not possible for training a TVG model directly (see Table~\ref{tab:pos}). After part-of-speech tagging, the average number of nouns and verbs for HowTo100M 30k videos are 73 and 36 respectively, since subtitles contain diverse objects and actions that can be used as candidate caption words or phrases for video moments, we reorganize nouns and verbs for each video moment by using a pre-trained vision-language model. 

\noindent\textbf{Evaluation datasets.} Two standard TVG datasets are evaluated: 1) \textit{ActivityNet Captions}~\cite{krishna2017dense}: ActivityNet Captions contains 20k videos amounting to 849 video hours with 100k total descriptions. On average, each of the 20K videos in ActivityNet Captions contains 3.65 temporally localized sentences, resulting in a total of 100k sentences.; 2) \textit{Charades-STA}~\cite{gao2017tall}: Charades-STA contains 9848 videos with an average duration of 30 seconds, annotated into 12,404 clip-sentence pairs for training set and 3,720 clip-sentence pairs for testing set.

\noindent\textbf{Implementation details.} The pre-training dataset HowTo100M is decoded at 2 fps, then 500 frames are uniformly sampled from overall extracted frames for generating captioned moments with CMG module. 
For video moment generation, we take 4 clusters for K-means clustering in practice, for moment caption selection,  we take top 5 nouns and top 3 verbs from candidate words to construct a caption for each moment.

We utilize CLIP ViT-B/32 backbone for CMG module and TVGNet. For a fair comparison with existing methods~\cite{nam2021zero} and ~\cite{wang2022prompt}, we extract I3D features pre-trained on Kinetics-400 dataset and C3D features pre-trained on Sports-1M dataset for pretraining dataset HowTo100M. Since the dimension of features extracted from C3D model is 4,096, we reduce the feature dimension to 500 with incremental PCA so as to keep it the same with downstream dataset features.
We use Adam optimizer with a fixed learning rate of 0.0004 for pre-training TVGNet with the vision encoder fixed. 128 frames are uniformly sampled from all datasets for pre-training and inference.

\noindent\textbf{Evaluation Metric}
Following previous works of TVG~\cite{nam2021zero}~\cite{mun2020local}, we use the ratio of recalled time
intervals whose tIoU (temporal Intersection over Union) with the ground truth is
larger than multiple thresholds (R@tIoU), e.g.,
$\text{tIoU}\in\{0.3, 0.5, 0.7\}$, as well as mean Intersection over Union (mIoU) metric for evaluation.

\subsection{Main results}

Tab.~\ref{tab:eval_charades} and Tab.~\ref{tab:eval_anet} show the results and comparisons with existing TVG methods on Charades-STA and ActivityNet Captions respectively. From both tables, it is not surprising to see that fully-supervised and some weakly-supervised methods surpass zero-shot counterparts since they enjoy more supervision, i.e., human-labeled ground-truth timestamps or captions. 
We compare AutoTVG with other weakly-supervised counterparts by applying video moment generation method to Charades-STA and take the moment that has the maximum similarity with its caption as the final prediction. We observe that AutoTVG is able to achieve highly comparable results with CTF~\cite{2020Look} and WSRA~\cite{fang2020weak}, and even surpasses SCN~\cite{lin2020weakly} by 4.2\% at R@0.5 and CTF~\cite{2020Look} by 1.1\% at mIoU, which shows our video moment generation module provides reliable candidate moments. Thanks to the strong alignment between vision and language features from CLIP, we can obtain decent weakly-supervised results under a simple non-parametric matching strategy.

\begin{table}[h]
    \centering
    \setlength{\tabcolsep}{3pt}
    \begin{tabular}{ccccc}
    \toprule
    \textbf{Method} & \textbf{Feature} & \textbf{R@0.5} & \textbf{R@0.7} & \textbf{mIoU} \\
    \midrule[.1pt]
    \multicolumn{5}{c}{\textit{Fully-supervised}}\\
     MLVI~\cite{xu2019multilevel} & C3D & 35.60 & 15.80 & - \\
     LGI~\cite{mun2020local} & I3D & 59.46 & 35.48 & 51.38 \\
     BPNet~\cite{xiao2021boundary} & I3D  & 50.75 & 31.64 & 46.34 \\
    \midrule[.1pt]
    \multicolumn{5}{c}{\textit{Weakly-supervised}}\\
     SCN~\cite{lin2020weakly} & C3D  & 23.58 & 9.97 & - \\
     CTF~\cite{2020Look}    &  C3D  &  27.30  &  12.90  &  27.30  \\
     WSRA~\cite{fang2020weak}  & C3D & 31.20   &   11.01  &  31.00   \\
     AutoTVG & CLIP  & 27.78 &  11.27  & 28.40  \\
    \midrule[.1pt]
    \multicolumn{5}{c}{\textit{Unsupervised}}\\
     DSCNet~\cite{liu2022unsupervised} & C3D & 28.73 & 14.67 & -  \\
    \midrule[.1pt]
    \multicolumn{5}{c}{\textit{Zero-shot}}\\
     \rowcolor{gray!20}
     PSVL~\cite{nam2021zero} & I3D & 31.29 & 14.17 & 31.24 \\
     \rowcolor{gray!20}
     PZVMR~\cite{wang2022prompt} & I3D & 33.21 & 18.51 & 32.62 \\
     \rowcolor{gray!20}
     PZVMR w/o $L_o$~\cite{wang2022prompt} & I3D & 29.85 & 17.14 & 29.53 \\
     UniVTG~\cite{lin2023univtg} &  CLIP & 25.22 & 10.03 & 27.12  \\
     AutoTVG & CLIP  &	26.21	& 11.86	& 25.39 \\
     \textbf{AutoTVG} & \textbf{I3D}  & \textbf{30.68} & \textbf{17.42} & \textbf{29.23}  \\
    \bottomrule
    \end{tabular}
    \vspace{1mm}
    \caption{Evaluation results on Charades-STA. AutoTVG shows superior performance in zero-shot setting and even outperforms some weakly-supervised methods. Methods in gray background conduct pseudo labeling and fine-tuning using Charades-STA.}
    \label{tab:eval_charades}
\end{table}

\begin{table}[h]
    \centering
    \begin{tabular}{cccc}
    \toprule
    \textbf{Method} & \textbf{Feature} & \textbf{R@0.3} & \textbf{R@0.5} \\
    \midrule[.1pt]
    \multicolumn{4}{c}{\textit{Fully-supervised}}\\
     MLVI~\cite{xu2019multilevel} & C3D & 45.30 & 27.70  \\
     ABLR~\cite{yuan2019find} & C3D & 55.67 & 36.79  \\
     LGI~\cite{mun2020local} & C3D & 58.52 & 41.51  \\
    \midrule[.1pt]
    \multicolumn{4}{c}{\textit{Weakly-supervised}}\\
     WSLLN~\cite{gao2019wslln} & C3D & 42.80 & 22.70  \\
     CTF~\cite{2020Look} & C3D  & 44.30 & 23.60 \\
    \midrule[.1pt]
    \multicolumn{4}{c}{\textit{Unsupervised}}\\
     DSCNet~\cite{liu2022unsupervised} & C3D & 47.29 & 28.16 \\
    \midrule[.1pt]
    \multicolumn{4}{c}{\textit{Zero-shot}}\\
     \rowcolor{gray!20}
     PSVL~\cite{nam2021zero} & C3D & 44.74 &  30.08  \\
     \rowcolor{gray!20}
     PZVMR~\cite{wang2022prompt} & C3D & 45.73 & 31.26  \\
     \rowcolor{gray!20}
     PZVMR w/o $L_o$ ~\cite{wang2022prompt} & C3D & 43.54 & 29.35  \\
     AutoTVG & CLIP & 36.16 & 15.60  \\
     \textbf{AutoTVG} &  \textbf{C3D}  &  \textbf{43.03}  &  \textbf{25.46} \\
     
    \bottomrule
    \end{tabular}
    \vspace{1mm}
    
    \caption{Evaluation results on ActivityNet Captions. AutoTVG shows superior performance in zero-shot setting. Methods in gray background conduct pseudo labeling and fine-tuning using ActivityNet Captions.}
    \label{tab:eval_anet}
\end{table}

Compared to zero-shot counterparts PSVL~\cite{nam2021zero} and PZVMR~\cite{wang2022prompt} which directly generate captioned moments on Charades-STA or ActivityNet Captions (i.e., the training and testing datasets are in-distribution), however, AutoTVG still achieves competitive results even though the pre-training dataset is out-of-distribution. 
On Charades-STA dataset, AutoTVG achieves 30.68\% at R@0.5 which is highly comparable to PSVL's 31.29\%, for R@0.7, AutoTVG achieves 17.42\% which surpasses PSVL by 3.25\%, and is comparable to PZVMR. We attribute this improvement mainly to the diversity of words in captions, because for both PSVL and PZVMR, the nouns are obtained from an off-the-shelf object detection model which would suffer from limited nouns defined by object detection datasets. While AutoTVG generates captions from automatically generated annotations (e.g., ASR) which is much more diverse in both nouns and verbs compared to object detection (see supplementary for details). As the common sense for pre-training, large-scale data provides rich knowledge to model learning so that it can bring benefit to out-of-distribution test. With more diverse action and object instances from HowTo100M pre-training dataset, the pre-trained TVG model is capable of zero-shot test on downstream datasets.
It is worth highlighting that our method achieves better results on R@0.5 and R@0.7 with only 30k annotations compared to UniVTG~\cite{lin2023univtg}, which requires 4.2M temporal annotations for grounding pre-training. This indicates that the proposed CMG can generate high-quality annotations, leading to a decent generalizability of AutoTVG.

On ActivityNet Captions dataset, we observe that AutoTVG is slightly lower on R@0.3 and worse on R@0.5 compared to zero-shot methods, we argue that the performance drop is mainly from the data bias between different datasets. Some hyper-parameters for captioned moment generation are essential for performance, such as the number of k-means clusters that would affect boundary precision, and the number of selected nouns and verbs that would affect moment and caption alignment during the pre-training process. As indicated in~\cite{yuan2021closer}, data bias would result in poor performance in out-of-distribution test, since we choose hyper-parameters and conduct ablation on Charades-STA dataset, the generated captioned moments are biased on Charades-STA dataset. We leave it a future work and expect there to be more effort into it.

\subsection{Ablation Studies}

All the ablation studies below are tested on Charades-STA dataset, with CLIP ViT-B-32 backbone for encoding video and text features if not otherwise specified.

\noindent\textbf{Video Moment Generation.}
\begin{table}
    \centering
    \setlength{\tabcolsep}{2pt}
    \begin{tabular}{ccccc}
    \toprule

    \begin{tabular}[c]{@{}l@{}}\textbf{Candidate}\\ \textbf{Generation}\end{tabular} & \begin{tabular}[c]{@{}l@{}}\textbf{Candidate}\\ \textbf{Selection}\end{tabular} & \textbf{R@0.3} & \textbf{R@0.5} & \textbf{R@0.7} \\
    \midrule[.1pt]
    \multicolumn{5}{c}{\textit{Without Pre-training}} \\
    Sliding Window & Max similarity & 25.45  & 5.29 & 0.75 \\
    Brute Force & Max similarity & 37.23 &	16.77 & 6.36 \\
    K-means & Max similarity & \textbf{45.81} & \textbf{27.78} &	\textbf{11.27} \\
    \rowcolor{gray!20}
    K-means & Perfect boundary &  45.28  &  27.15  & 12.60 \\
    \rowcolor{gray!20}
    K-means & Perfect alignment &  98.73 &  63.48	&  23.68    \\
    \midrule[.1pt]
    \multicolumn{5}{c}{\textit{With Pre-training}}\\
    K-means & Random & \textbf{37.07} & \textbf{26.21} & 11.86 \\
    K-means & Longest & 36.72 & 25.78 & \textbf{13.04} \\
    K-means & Distinct & 34.41 & 20.35 & 7.66 \\
    \bottomrule
    \end{tabular}
    \vspace{1mm}
    
    \caption{Ablation for video moment generation and moment caption selection in CMG module. Rows in gray background are upper bound analyses of boundary and alignment.}
    \label{tab:ablation_moment}
\end{table}
Table~\ref{tab:ablation_moment} shows the ablation study of video moment generation process.
We evaluate video moment generation methods under \textit{Without pre-training} setting, i.e., directly apply moment generation to Charades-STA dataset without pre-training TVG model, the optimal moment is selected by the maximum similarity between candidate moments and original video captions. Three candidate generation methods are evaluated: 1) Sliding Window: we set a fixed window size of 8 and stride of 4; 2) Brute Force: we construct a 2D-map to represent all possible candidate moments and calculate the similarity with the caption for each moment; 3) K-means: we concatenate frame index to frame features and then take K-means to cluster frame features. We observe that K-means surpasses other methods by a clear margin which indicates clustering frames based on contextual visual clues is a superior strategy.

For further analyzing the reliability of candidate generation and candidate selection method, we conduct Perfect boundary and Perfect alignment candidate selection strategies: 1) Perfect boundary: assume that K-means can generate one moment which is the same as the ground truth moment, we add ground truth moment to candidate moments then select the moment with maximum similarity with the caption; 2) Perfect alignment: assume that the candidate moment which has the maximum tIoU with ground truth moment also has the maximum similarity with the caption, i.e. the caption and the moment has a strong alignment, we take one moment from candidates which have the maximum tIoU with ground truth moment. 

We observe that K-means with Perfect boundary has similar results with K-means with Max similarity (R@0.3: 45.28\% vs. 45.81\%), which indicates that even if the model can generate perfect boundary, however, it can not achieve higher performance without great alignment ability. Also, K-means with Perfect alignment has much better performance than K-means with Max similarity (R@0.3: 98.73\% vs. 45.81\%) which means even if the boundary is not precise enough, the model can still predict satisfactory moments with strong alignment ability. Our framework brings the merit of CLIP for moment and caption alignment so that the generated captioned moments are reliable for pre-training a TVG model.


\noindent\textbf{Moment Candidates Selection.}
We evaluate the moment candidates selection strategies under \textit{With pre-training} setting, i.e. generate captioned moments and then take the captioned moments to pre-train TVGNet, which is afterward used to evaluate on Charades-STA dataset.
Table~\ref{tab:ablation_moment} shows the ablation for moment candidates selection. We test three candidate selection methods including 1) Random: random sample one moment from candidates; 2) Longest: take the longest moment from candidates; 3) Distinct: take the moment whose feature has the largest distance with average moment features. We observe that Distinct strategy obtains the lowest performance in R@0.3 and R@0.5 while Random strategy obtains the best results, maybe random selection reduces the bias of moment length for model training so that model can adapt to downstream datasets with various lengths of videos. Since the pre-training videos are untrimmed and have noisy backgrounds, e.g., videos may contain video opening and closing credits, Distinct strategy is likely to bring meaningless noise.

\begin{table}[h]
    \centering
    \setlength{\tabcolsep}{2pt}
    \begin{tabular}{l*3{c}}
    \toprule
    \textbf{Caption Types} & \textbf{R@0.3} & \textbf{R@0.5} & \textbf{R@0.7} \\
    \midrule
    None &  failed &  failed  &  failed  \\
    CMG w/o PoS & 29.65 & 17.85	& 5.81  \\
    CMG w/ PoS &  \textbf{37.07} & \textbf{26.21} & \textbf{11.86}   \\
    \bottomrule
    \end{tabular}
    \vspace{1mm}
    
    \caption{Ablation for Part-of-Speech Tagging}
    \label{tab:pos}
\end{table}

\noindent\textbf{Effectiveness of Part-of-Speech Tagging.}
We evaluate the effectiveness of Part-of-Speech Tagging including 1) None: directly apply automatically generated subtitles and timestamps of HowTo100M to pre-train TVG model; 2) CMG w/o PoS: implement CMG without Part-of-Speech Tagging; 3) CMG w/ PoS: implement CMG with Part-of-Speech Tagging. From Table~\ref{tab:pos} we can see automatically generated raw annotations of HowTo100M are too noisy to train a TVG model. Part-of-Speech tagging plays an important role in moment caption selection by reducing the large number of noisy words in subtitles.

\begin{figure*}[h]
    \centering
    \includegraphics[width=0.96\linewidth]{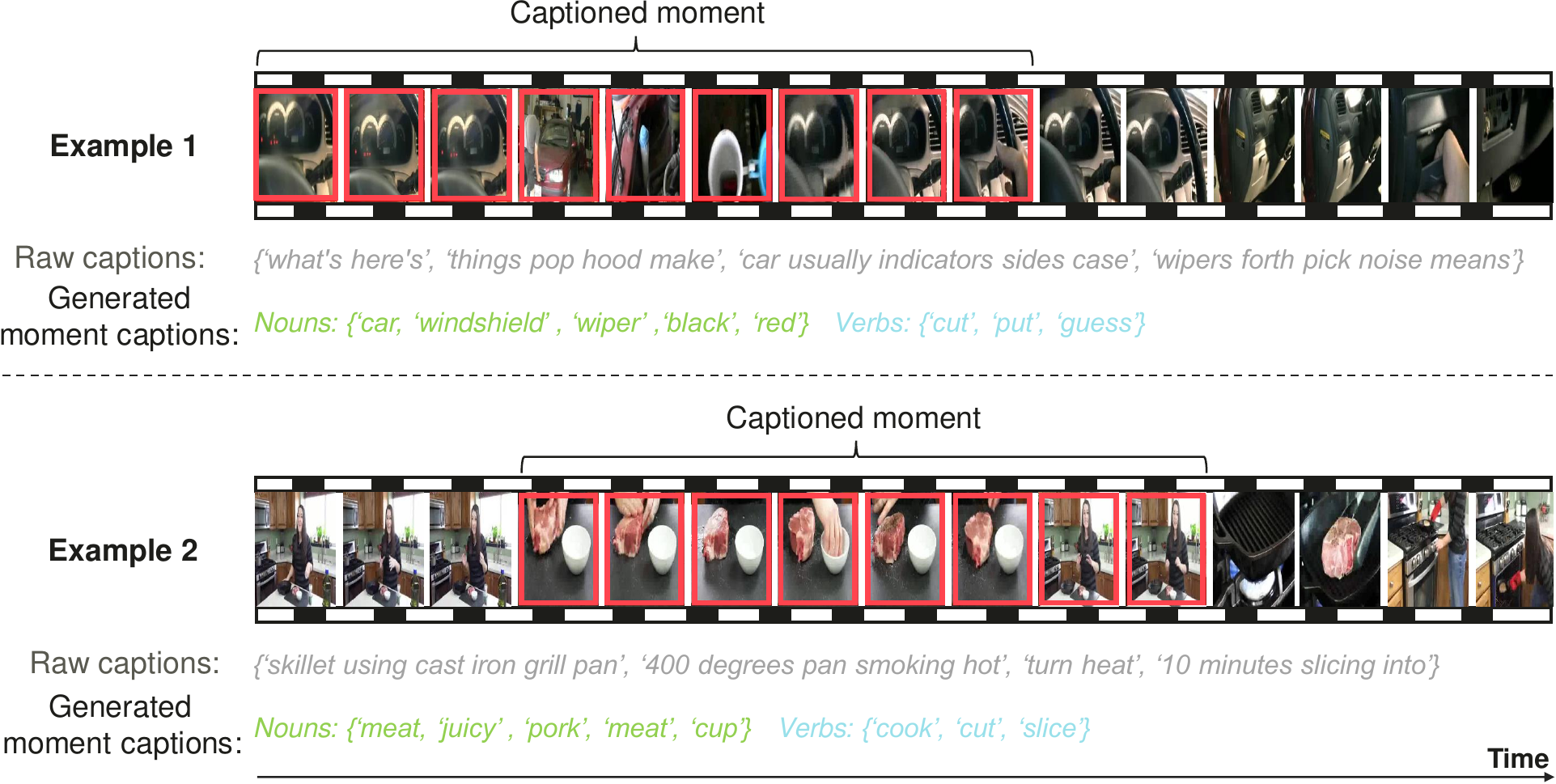}
    \caption{Visualization of the generated captioned moments and raw video captions.}
    \label{fig:visualization}
\end{figure*}


\begin{table}[h]
    \centering
    \begin{tabular}{c*4{c}}
    \toprule
    \textbf{\#Noun} & \textbf{\#Verb} & \textbf{R@0.3} & \textbf{R@0.5} & \textbf{R@0.7} \\
    \midrule
    5 & 0 & \textbf{38.31}	& 24.41	& 10.91  \\
    5 & 3 & 37.07 & \textbf{26.21} & \textbf{11.86}  \\
    5 & 3 (random) &  34.38  &  21.45  &  9.17  \\
    10 & 6 & 30.62	& 19.00 & 8.79 \\
    scale & scale & 34.89 & 21.13	& 9.36 \\
    \bottomrule
    \end{tabular}
    \vspace{1mm}
    
    \caption{Ablation for nouns and verbs.}
    \label{tab:ablation_word}
\end{table}

\noindent\textbf{Effects of Nouns and Verbs.}
(1) Number of nouns and verbs: We evaluate the effects of different numbers of nouns and verbs including fixed and scale numbers (i.e., the number of nouns and verbs are in proportion to moment's length). Especially, we evaluate the effect of verbs that are significant for identifying actions and activities. As we can see in Table~\ref{tab:ablation_word}, removing verbs will result in 1.8\% and 0.95\% performance drop to R@5 and R@7, random samples 3 verbs will result in 4.76\% and 2.69\% performance drop to R@5 and R@7.

(2) Diversity of nouns and verbs:
We compare AutoTVG to zero-shot counterparts PSVL~\cite{nam2021zero} and PZVMR~\cite{wang2022prompt} which generate nouns via off-the-shelf object detection model, to show the diversity of nouns and verbs. To be specific, PSVL uses a Faster-RCNN~\cite{ren2015faster} model pre-trained on Visual Genome~\cite{krishna2017visual} dataset with 1,600 classes, PZVMR uses a Faster-RCNN model pre-trained on COCO dataset~\cite{lin2014microsoft} and Pascal VOC~\cite{everingham2009pascal} dataset.
AutoTVG generates captions from automatically generated subtitles (e.g., ASR) which are much more diverse in both nouns and verbs.

\begin{table}[h]
    \centering
    \begin{tabular}{cccc}
    \toprule
      Method   & Dataset & \#Nouns & \#Verbs \\
    \midrule
      PSVL & Visual Genome  &  1,600 &   0 \\ 
      PZVMR & COCO & 80  &   0  \\
      PZVMR & Pascal VOC & 20 &   0 \\
    \midrule
      AutoTVG & HT100M 10k & 557,878  & 12,350  \\
      AutoTVG & HT100M 20k & 1,069,097  &  17,061 \\
      AutoTVG & HT100M 30k &  1,591,031 &  20,617 \\
    \bottomrule
    \end{tabular}
    \vspace{1mm}
    \caption{Number of nouns and verbs generated by object detection models and subtitles.}
    \label{tab:diverse}
\end{table}

As shown in Table~\ref{tab:diverse}, we analyze the number of nouns and verbs for different HowTo100M subsets, and we count the data after Part-of-Speech tagging preprocessing. As we can see, the nouns and verbs from subtitles are much more diverse than those from pre-trained object detection models.

\begin{table}[h]
    \centering
    \begin{tabular}{c*4{c}}
    \toprule
      Text Encoder   &  R@0.5  & R@0.7  & mIoU  \\
    \midrule
      CLIP    & \textbf{30.68} &  \textbf{17.42} & 29.20 \\
      PSVL~\cite{nam2021zero}   &  20.40 & 6.64 & \textbf{29.29} \\
    \bottomrule
    \end{tabular}
    \vspace{1mm}
    
    \caption{Comparison between CLIP text encoder and text encoder from PSVL, results are tested on I3D feature.}
    \label{tab:text_encode}
\end{table}

\noindent\textbf{Effects of TVGNet Text Encoder.}
As mentioned in method section, we think directly adapting previous text encoding to TVGNet is problematic, because the vocabulary dictionary is built based on word frequency of training data, which is not applicable to out-of-distribution test.
To avoid text encoding bias to the word of pre-training dataset, we take CLIP~\cite{radford2021learning} text encoder with a large vocabulary size of 49,152 to extract text features.
Table~\ref{tab:text_encode} shows the results of CLIP text encoder and original text encoder in PSVL~\cite{nam2021zero}, we observe that the text encoder from CLIP surpasses that from PSVL by a clear margin.

\begin{table}[h]
    \centering
    \begin{tabular}{c*3{c}}
    \toprule
      Clustering Strategy   &      R@0.3   &  R@0.5  & R@0.7   \\
    \midrule
       w/o index  &  30.02 & 18.24 &  8.41 \\
       w/ index   &  \textbf{45.81} & \textbf{27.78} &  \textbf{11.27}   \\
    \bottomrule
    \end{tabular}
    \vspace{1mm}
    
    \caption{Effectiveness of concatenated frame indices for K-means.}
    \label{tab:index}
\end{table}

\begin{table}[h]
    \centering
    \begin{tabular}{c*3{c}}
    \hline
    \textbf{\#Clusters} & \textbf{R@0.3} & \textbf{R@0.5} & \textbf{R@0.7} \\
    \hline
    3 & \textbf{40.43}	& 22.63	& 8.47 \\
    4 & 37.07  & \textbf{26.21} & \textbf{11.86} \\
    5 & 28.20	& 16.10	& 6.37 \\
    \hline
    \end{tabular}
    \vspace{1mm}
    \caption{Ablation for the number of K-means clusters.}
    \label{tab:ablation_cluster}
\end{table}

\noindent
\textbf{Effects of K-means Clusters.}
For encouraging K-means to cluster adjacent frames as
a candidate moment, we concatenate frame features with their frame indices, Table~\ref{tab:index} shows the effectiveness of frame indices for K-means under \textit{Without Pre-training} setting. We evaluate the number of clusters for K-means algorithm in Table~\ref{tab:ablation_cluster}, in practice, we take cluster number as 4 which achieves the highest performance.


\begin{table}[h]
    \centering
    \begin{tabular}{cccc}
    \toprule
    \textbf{\# Pre-train Data} & \textbf{R@0.5} & \textbf{R@0.7} & \textbf{mIoU} \\
    \midrule
      15k &  28.66 & 14.33 & 28.01   \\
      30k &  30.68 & 17.42 & 29.23    \\
    \bottomrule
    \end{tabular}
    \vspace{1mm}
    
    \caption{Incremental results on Charades-STA dataset tested with I3D feature.}
    \label{tab:incremental}
\end{table}

\begin{table}[h]
    \centering
    \setlength{\tabcolsep}{2.5pt}
    \begin{tabular}{cccccc}
    \toprule
     \textbf{Method} & \textbf{ Train Dataset} & \textbf{\# Videos}  & \textbf{R@0.5} & \textbf{R@0.7} & \textbf{mIoU} \\
    \midrule
      PSVL [28]  &  Charades-STA  & 12K & 	28.17  &  14.92  &  \textbf{30.24}  \\ 
      CMG   &  HowTo100M  &  12K  &    29.60   &  15.72  &  29.22  \\
      CMG   &  HowTo100M &  30K  &  \textbf{30.68}  &  \textbf{17.42}  &  29.23  \\
    \bottomrule
    \end{tabular}
    \vspace{1mm}
    \caption{In-distribution zero-shot testing results tested on Charades-STA, I3D feature is evaluated.}
    \label{tab:IID-test}
\end{table}

\noindent
\textbf{Effects of Training Data Amount}
To further illustrate how data amount affects the generalization ability of TVGNet, we conduct experiments with 15k videos randomly split from the 30k subset. Experimental results in Table~\ref{tab:incremental} show that the out-of-distribution test can be improved by increasing training data.
We use open-source Charades-STA pseudo labels generated by PSVL~\cite{nam2021zero} to train our TVGNet and test on Charades-STA, to further compare in-distribution and out-of-distribution results under different training data amount. Table~\ref{tab:IID-test} shows that when taking 12K training data, CMG \& HowTo100M is comparable with PSVL \& Charades-STA. With the training data increasing, i.e. 30K, CMG \& HowTo100M can achieve better results.

\subsection{Visualization}
Figure~\ref{fig:visualization} shows two examples generated from the proposed CMG module. As we can see, the proposed CMG can accurately match video moments with corresponding nouns and verbs. For instance, example 2 shows that the nouns of \textit{``meat'', ``juicy'', ``pork'', ``meat'', ``cup''} and the verbs of \textit{``cook'', ``cut'', ``slice''} are more appropriate than raw captions which indicates CMG module can produce reliable captioned moments.

%% file: conclusion.tex
\vspace{-3mm}
\section{Conclusion}
\label{sec:conclusion}
\vspace{-1mm}


Temporal video grounding (TVG) with limited supervision is a challenging task that has attracted much attention from both academia and industry. Inspired by the great success of pre-training works on classification and retrieval tasks, researchers also design pretext tasks for TVG following ``pre-training + fine-tunining'' paradigm. However, this paradigm has two obvious drawbacks, one is the lack of temporal modeling and fine-grained alignment, and the other is the large gap between pretext and downstream task. To this end, we propose AutoTVG to avoid the problem and achieve decent performance on out-of-distribution test with automatically annotated pre-train data. We expect there to be more efforts to reduce data bias and break the limitations of CMG in the future.

%% file: ht100m.tex
\section{Analysis of pre-training dataset}
\label{sec:ht100m}

\subsection{Task Categories of Subset}
We randomly sampled a subset of 30K videos from HowTo100M dataset, now we provide more details about the subset. Each video from HowTo100M has a highest level task category from WikiHow and a second highest level task category from WikiHow, the subset contains almost all kinds of tasks in the whole set, see Table~\ref{tab:ht30k}.

\begin{table}[h]
    \centering
    \setlength{\tabcolsep}{2.5pt}
    \begin{tabular}{c*4{c}}
    \toprule
      \textbf{Dataset} & \textbf{\#Main task}   &   \textbf{\#Sub-task}  \\
    \midrule
     HowTo100M  &  20  &  143  \\
     HowTo100M subset & 20  &  138 \\
    \bottomrule
    \end{tabular}
    \vspace{1mm}
    \caption{Statistics of the 30K HowTo100M subset, the subset covers almost all kinds of tasks in the whole set.}
    \label{tab:ht30k}
\end{table}

\subsection{Diversity of Nouns and Verbs}

AutoTVG generates captions from automatically generated subtitles (e.g., ASR) which are much more diverse in both nouns and verbs. We provide more qualitative results to illustrate it.

\begin{figure*}[b]
    \centering
    \includegraphics[width=0.8\linewidth]{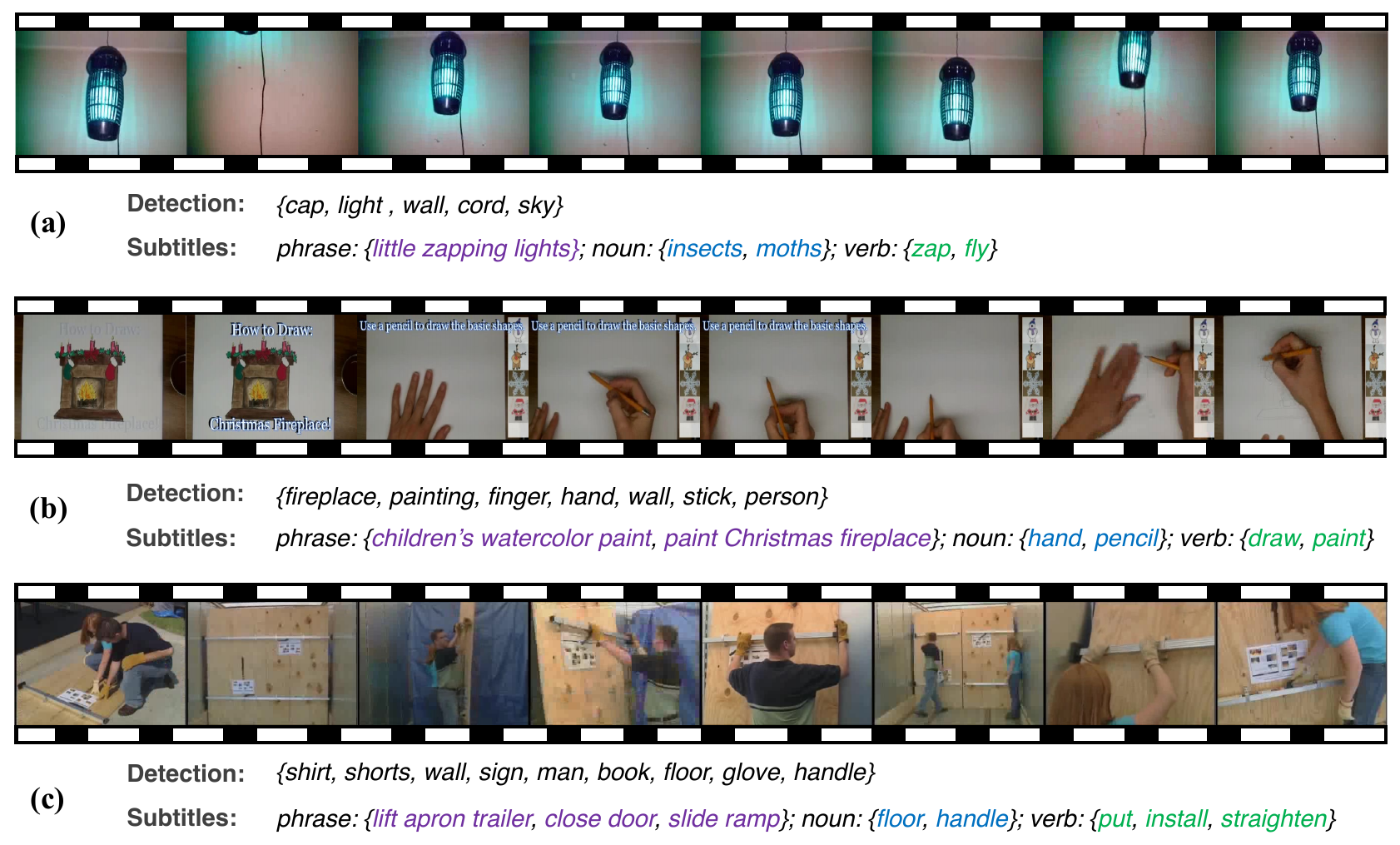}
    \caption{Comparison of words from object detection model and words from automatically generated subtitles. Subtitles are superior in describing very tiny objects (a), detailed concepts (b), and visually incomprehensible actions (c).}
    \label{fig:case}
\end{figure*}

We show some cases to explain why automatically generated subtitles are superior to detected words from object detection models~\cite{nam2021zero, wang2022prompt} in detail. For both methods, we first generate a candidate moment for each video with Video Moment Generation in CMG, then we apply Faster-RCNN model which was pre-trained on Visual Genome with ResNet-101 as the backbone, to detect object for each frame of the moment. Meanwhile, we use Moment Caption Selection in CMG to select nouns and verbs from subtitles. After that, we compare the words from the detection model to subtitles as shown in Figure~\ref{fig:case}.

The automatically generated subtitles are superior to objects detected from models, especially in describing tiny objects, detailed concepts, and visually incomprehensible actions. 
From Figure~\ref{fig:case} (a), we observe that the insects flying around the light and moths killed inside the light are too tiny to be detected by the object detection model, but are presented in subtitles, which indicates subtitles can be employed to compensate for object detection by tackling tiny objects.
From Figure~\ref{fig:case} (b), we can see the phrases from subtitles contain more detailed concepts, i.e., \textit{``children's watercolor paint''} \textit{vs.} \textit{``painting''} and \textit{``paint Christmas fireplace''} \textit{vs.} \textit{``fireplace''}, besides,\textit{``pencil''} is more accurate than \textit{``stick''}.
Figure~\ref{fig:case} (c) shows two people explaining how to properly load a moving trailer, the complex actions are incomprehensible if we only watch the video without any captions. The object detection model only provides objects appearing in this video that are insufficient to describe the actions, whereas verbs such as \textit{``install''} and \textit{``straighten''} combined with \textit{``lift apron trailer''} and \textit{``slide ramp''} from subtitles can facilitate understanding the visually incomprehensible actions.

As we can see, automatically generated subtitles extend visual concepts with diverse phrases, nouns and verbs, therefore the pre-trained model is semantic-rich and can generalize to downstream datasets.

%% file: limitation.tex
\section{Limitations of CMG}
\label{sec:limitation}

\begin{figure*}[t]
    \centering
    \includegraphics[width=0.8\linewidth]{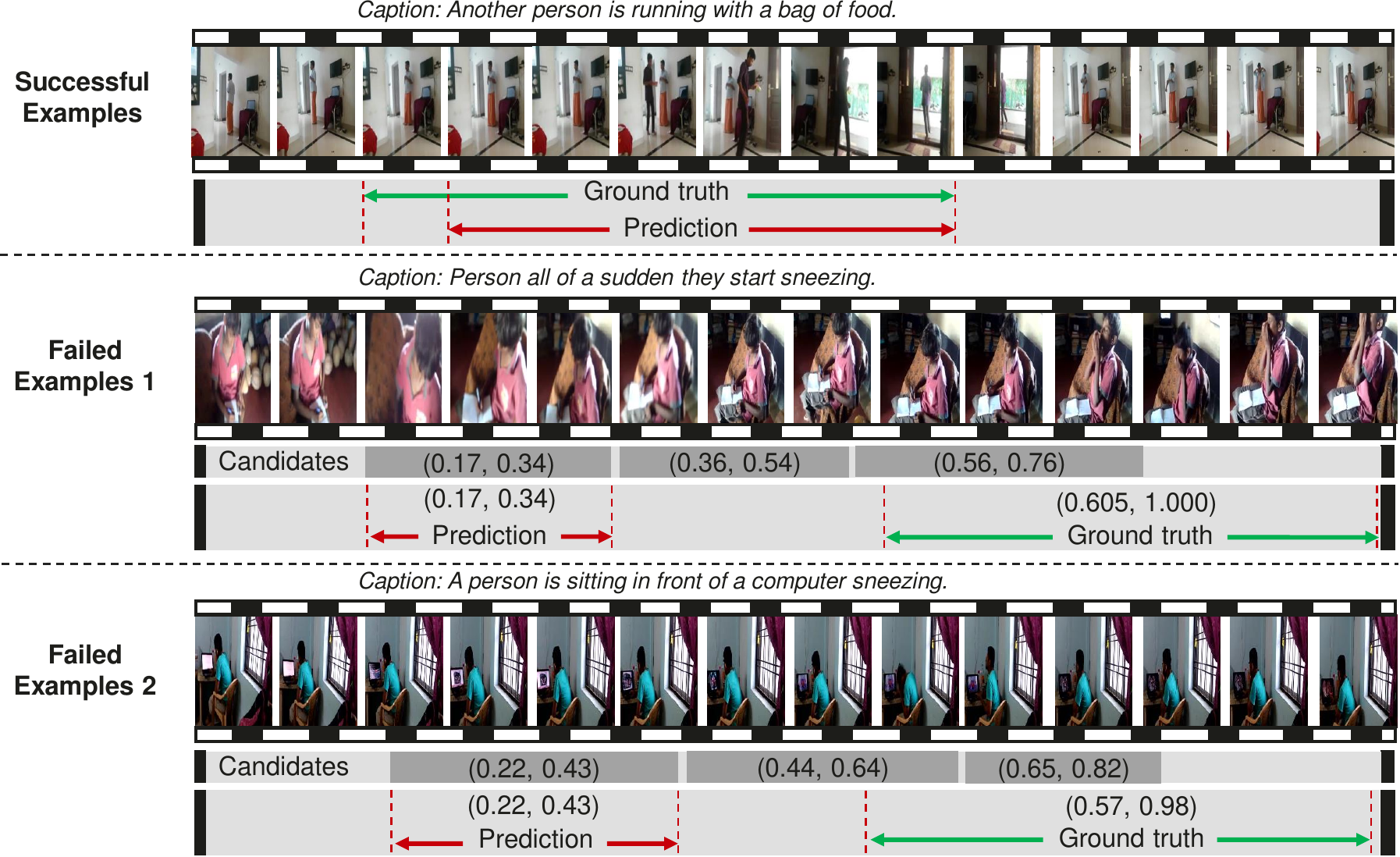}
    \caption{Successful and failed cases from Charades-STA. CLIP wrongly aligns a moment that has high similarity with the caption while low tIoU with the ground truth moment (best viewed in color).}
    \label{fig:limitation}
\end{figure*}

This section shows some typical cases of Charades-STA to illustrate the limitations of CMG module. As aforementioned in ablation studies of CMG, the visual-language alignment ability of CLIP provides satisfactory results but will fail in some cases. From Table~\ref{fig:limitation} we find that \textit{"sneezing"} is a challenging class for grounding. We analyze their candidate moments generated by video moment generation module and find that CLIP wrongly aligns a moment that has high similarity with the caption while low tIoU with the ground truth moment, perhaps the image-text pre-trained CLIP model has some limitations on video-text alignment without video pre-training so that some videos containing actions are failed to be matched to verbs. For further convincing our assumption, we also conduct experiments on CLIP model pre-trained on more image-text paired data and video-text paired data including WebVid-2M~\cite{bain2021frozen} and CC3M~\cite{sharma2018conceptual}. From Table~\ref{tab:limitation} we observe that more video-text paired pre-training data can benefit CMG module to align candidate moments and captions more precisely.

\begin{table}[h]
    \centering
    \begin{tabular}{c*3{c}}
    \toprule
      \textbf{Pre-training datasets}   &  \textbf{R@0.3} & \textbf{R@0.5}  & \textbf{R@0.7}  \\
    \midrule
     \rowcolor{gray!20}
      CLIP~\cite{radford2021learning}  &  45.81 & 27.78 & 11.27    \\
      WebVid-200K  &  48.27   &  28.99  & 12.08   \\
      CC3M     &   49.02   &   29.83 &   11.63 \\
      WebVid-2M+CC3M~\cite{ge2022bridging}   & \textbf{50.31}  &  \textbf{30.66} & \textbf{12.66}  \\
    \bottomrule
    \end{tabular}
    \vspace{1mm}
    
    \caption{Results on CLIP model pre-trained with more datasets. The row in gray background is CLIP loaded with pre-trained model weights from~\cite{radford2021learning},  and other rows pre-train CLIP with more image-text (CC3M) or video-text (WebVid) paired datasets.}
    \label{tab:limitation}
\end{table}

%% file: camera-ready.bbl

\begin{thebibliography}{61}


\ifx \showCODEN    \undefined \def \showCODEN     #1{\unskip}     \fi
\ifx \showDOI      \undefined \def \showDOI       #1{#1}\fi
\ifx \showISBNx    \undefined \def \showISBNx     #1{\unskip}     \fi
\ifx \showISBNxiii \undefined \def \showISBNxiii  #1{\unskip}     \fi
\ifx \showISSN     \undefined \def \showISSN      #1{\unskip}     \fi
\ifx \showLCCN     \undefined \def \showLCCN      #1{\unskip}     \fi
\ifx \shownote     \undefined \def \shownote      #1{#1}          \fi
\ifx \showarticletitle \undefined \def \showarticletitle #1{#1}   \fi
\ifx \showURL      \undefined \def \showURL       {\relax}        \fi
\providecommand\bibfield[2]{#2}
\providecommand\bibinfo[2]{#2}
\providecommand\natexlab[1]{#1}
\providecommand\showeprint[2][]{arXiv:#2}

\bibitem[Akbari et~al\mbox{.}(2021)]%
        {akbari2021vatt}
\bibfield{author}{\bibinfo{person}{Hassan Akbari}, \bibinfo{person}{Liangzhe Yuan}, \bibinfo{person}{Rui Qian}, \bibinfo{person}{Wei-Hong Chuang}, \bibinfo{person}{Shih-Fu Chang}, \bibinfo{person}{Yin Cui}, {and} \bibinfo{person}{Boqing Gong}.} \bibinfo{year}{2021}\natexlab{}.
\newblock \showarticletitle{Vatt: Transformers for multimodal self-supervised learning from raw video, audio and text}.
\newblock \bibinfo{journal}{\emph{Advances in NIPS}}  \bibinfo{volume}{34} (\bibinfo{year}{2021}), \bibinfo{pages}{24206--24221}.
\newblock


\bibitem[Alwassel et~al\mbox{.}(2021)]%
        {alwassel2021tsp}
\bibfield{author}{\bibinfo{person}{Humam Alwassel}, \bibinfo{person}{Silvio Giancola}, {and} \bibinfo{person}{Bernard Ghanem}.} \bibinfo{year}{2021}\natexlab{}.
\newblock \showarticletitle{Tsp: Temporally-sensitive pretraining of video encoders for localization tasks}. In \bibinfo{booktitle}{\emph{ICCV}}. \bibinfo{pages}{3173--3183}.
\newblock


\bibitem[Anne~Hendricks et~al\mbox{.}(2017)]%
        {anne2017localizing}
\bibfield{author}{\bibinfo{person}{Lisa Anne~Hendricks}, \bibinfo{person}{Oliver Wang}, \bibinfo{person}{Eli Shechtman}, \bibinfo{person}{Josef Sivic}, \bibinfo{person}{Trevor Darrell}, {and} \bibinfo{person}{Bryan Russell}.} \bibinfo{year}{2017}\natexlab{}.
\newblock \showarticletitle{Localizing moments in video with natural language}. In \bibinfo{booktitle}{\emph{ICCV}}. \bibinfo{pages}{5803--5812}.
\newblock


\bibitem[Bain et~al\mbox{.}(2021)]%
        {bain2021frozen}
\bibfield{author}{\bibinfo{person}{Max Bain}, \bibinfo{person}{Arsha Nagrani}, \bibinfo{person}{G{\"u}l Varol}, {and} \bibinfo{person}{Andrew Zisserman}.} \bibinfo{year}{2021}\natexlab{}.
\newblock \showarticletitle{Frozen in time: A joint video and image encoder for end-to-end retrieval}. In \bibinfo{booktitle}{\emph{Proceedings of the IEEE/CVF International Conference on Computer Vision}}. \bibinfo{pages}{1728--1738}.
\newblock


\bibitem[Cao et~al\mbox{.}(2022)]%
        {cao2022locvtp}
\bibfield{author}{\bibinfo{person}{Meng Cao}, \bibinfo{person}{Tianyu Yang}, \bibinfo{person}{Junwu Weng}, \bibinfo{person}{Can Zhang}, \bibinfo{person}{Jue Wang}, {and} \bibinfo{person}{Yuexian Zou}.} \bibinfo{year}{2022}\natexlab{}.
\newblock \showarticletitle{Locvtp: Video-text pre-training for temporal localization}. In \bibinfo{booktitle}{\emph{ECCV}}. Springer, \bibinfo{pages}{38--56}.
\newblock


\bibitem[Chen et~al\mbox{.}(2023)]%
        {chen2023curriculum}
\bibfield{author}{\bibinfo{person}{Houlun Chen}, \bibinfo{person}{Xin Wang}, \bibinfo{person}{Xiaohan Lan}, \bibinfo{person}{Hong Chen}, \bibinfo{person}{Xuguang Duan}, \bibinfo{person}{Jia Jia}, {and} \bibinfo{person}{Wenwu Zhu}.} \bibinfo{year}{2023}\natexlab{}.
\newblock \showarticletitle{Curriculum-listener: Consistency-and complementarity-aware audio-enhanced temporal sentence grounding}. In \bibinfo{booktitle}{\emph{Proceedings of the 31st ACM International Conference on Multimedia}}. \bibinfo{pages}{3117--3128}.
\newblock


\bibitem[Chen et~al\mbox{.}(2020)]%
        {2020Look}
\bibfield{author}{\bibinfo{person}{Zhenfang Chen}, \bibinfo{person}{Lin Ma}, \bibinfo{person}{Wenhan Luo}, \bibinfo{person}{Peng Tang}, {and} \bibinfo{person}{Kwan Yee~K. Wong}.} \bibinfo{year}{2020}\natexlab{}.
\newblock \bibinfo{title}{Look closer to ground better: Weakly-supervised temporal grounding of sentence in video}.
\newblock
\newblock


\bibitem[Ding et~al\mbox{.}(2021)]%
        {ding2021support}
\bibfield{author}{\bibinfo{person}{Xinpeng Ding}, \bibinfo{person}{Nannan Wang}, \bibinfo{person}{Shiwei Zhang}, \bibinfo{person}{De Cheng}, \bibinfo{person}{Xiaomeng Li}, \bibinfo{person}{Ziyuan Huang}, \bibinfo{person}{Mingqian Tang}, {and} \bibinfo{person}{Xinbo Gao}.} \bibinfo{year}{2021}\natexlab{}.
\newblock \showarticletitle{Support-set based cross-supervision for video grounding}. In \bibinfo{booktitle}{\emph{ICCV}}. \bibinfo{pages}{11573--11582}.
\newblock


\bibitem[Duan et~al\mbox{.}(2018)]%
        {duan2018weakly}
\bibfield{author}{\bibinfo{person}{Xuguang Duan}, \bibinfo{person}{Wenbing Huang}, \bibinfo{person}{Chuang Gan}, \bibinfo{person}{Jingdong Wang}, \bibinfo{person}{Wenwu Zhu}, {and} \bibinfo{person}{Junzhou Huang}.} \bibinfo{year}{2018}\natexlab{}.
\newblock \showarticletitle{Weakly supervised dense event captioning in videos}.
\newblock \bibinfo{journal}{\emph{Advances in NIPS}}  \bibinfo{volume}{31} (\bibinfo{year}{2018}).
\newblock


\bibitem[Everingham et~al\mbox{.}(2009)]%
        {everingham2009pascal}
\bibfield{author}{\bibinfo{person}{Mark Everingham}, \bibinfo{person}{Luc Van~Gool}, \bibinfo{person}{Christopher~KI Williams}, \bibinfo{person}{John Winn}, {and} \bibinfo{person}{Andrew Zisserman}.} \bibinfo{year}{2009}\natexlab{}.
\newblock \showarticletitle{The pascal visual object classes (voc) challenge}.
\newblock \bibinfo{journal}{\emph{International journal of computer vision}}  \bibinfo{volume}{88} (\bibinfo{year}{2009}), \bibinfo{pages}{303--308}.
\newblock


\bibitem[Fang et~al\mbox{.}(2020)]%
        {fang2020weak}
\bibfield{author}{\bibinfo{person}{Zhiyuan Fang}, \bibinfo{person}{Shu Kong}, \bibinfo{person}{Zhe Wang}, \bibinfo{person}{Charless Fowlkes}, {and} \bibinfo{person}{Yezhou Yang}.} \bibinfo{year}{2020}\natexlab{}.
\newblock \showarticletitle{Weak supervision and referring attention for temporal-textual association learning}.
\newblock \bibinfo{journal}{\emph{arXiv preprint arXiv:2006.11747}} (\bibinfo{year}{2020}).
\newblock


\bibitem[Gao et~al\mbox{.}(2017)]%
        {gao2017tall}
\bibfield{author}{\bibinfo{person}{Jiyang Gao}, \bibinfo{person}{Chen Sun}, \bibinfo{person}{Zhenheng Yang}, {and} \bibinfo{person}{Ram Nevatia}.} \bibinfo{year}{2017}\natexlab{}.
\newblock \showarticletitle{Tall: Temporal activity localization via language query}. In \bibinfo{booktitle}{\emph{ICCV}}. \bibinfo{pages}{5267--5275}.
\newblock


\bibitem[Gao et~al\mbox{.}(2019)]%
        {gao2019wslln}
\bibfield{author}{\bibinfo{person}{Mingfei Gao}, \bibinfo{person}{Larry~S Davis}, \bibinfo{person}{Richard Socher}, {and} \bibinfo{person}{Caiming Xiong}.} \bibinfo{year}{2019}\natexlab{}.
\newblock \showarticletitle{Wslln: Weakly supervised natural language localization networks}.
\newblock \bibinfo{journal}{\emph{arXiv preprint arXiv:1909.00239}} (\bibinfo{year}{2019}).
\newblock


\bibitem[Gao et~al\mbox{.}(2020)]%
        {gao2020weakly}
\bibfield{author}{\bibinfo{person}{Mingfei Gao}, \bibinfo{person}{Richard Socher}, {and} \bibinfo{person}{Caiming Xiong}.} \bibinfo{year}{2020}\natexlab{}.
\newblock \bibinfo{title}{Weakly supervised natural language localization networks}.
\newblock
\newblock
\newblock
\shownote{US Patent App. 16/531,343}.


\bibitem[Ge et~al\mbox{.}(2022)]%
        {ge2022bridging}
\bibfield{author}{\bibinfo{person}{Yuying Ge}, \bibinfo{person}{Yixiao Ge}, \bibinfo{person}{Xihui Liu}, \bibinfo{person}{Dian Li}, \bibinfo{person}{Ying Shan}, \bibinfo{person}{Xiaohu Qie}, {and} \bibinfo{person}{Ping Luo}.} \bibinfo{year}{2022}\natexlab{}.
\newblock \showarticletitle{Bridging Video-Text Retrieval With Multiple Choice Questions}. In \bibinfo{booktitle}{\emph{Proceedings of the IEEE/CVF Conference on Computer Vision and Pattern Recognition}}. \bibinfo{pages}{16167--16176}.
\newblock


\bibitem[Hessel et~al\mbox{.}(2021)]%
        {hessel2021clipscore}
\bibfield{author}{\bibinfo{person}{Jack Hessel}, \bibinfo{person}{Ari Holtzman}, \bibinfo{person}{Maxwell Forbes}, \bibinfo{person}{Ronan~Le Bras}, {and} \bibinfo{person}{Yejin Choi}.} \bibinfo{year}{2021}\natexlab{}.
\newblock \showarticletitle{Clipscore: A reference-free evaluation metric for image captioning}.
\newblock \bibinfo{journal}{\emph{arXiv preprint arXiv:2104.08718}} (\bibinfo{year}{2021}).
\newblock


\bibitem[Honnibal et~al\mbox{.}(2020)]%
        {spacy}
\bibfield{author}{\bibinfo{person}{Matthew Honnibal}, \bibinfo{person}{Ines Montani}, \bibinfo{person}{Sofie~Van Landeghem}, {and} \bibinfo{person}{Adriane Boyd}.} \bibinfo{year}{2020}\natexlab{}.
\newblock \showarticletitle{spaCy: Industrial-strength Natural Language Processing in Python}.
\newblock


\bibitem[Huang et~al\mbox{.}(2021)]%
        {huang2021cross}
\bibfield{author}{\bibinfo{person}{Jiabo Huang}, \bibinfo{person}{Yang Liu}, \bibinfo{person}{Shaogang Gong}, {and} \bibinfo{person}{Hailin Jin}.} \bibinfo{year}{2021}\natexlab{}.
\newblock \showarticletitle{Cross-sentence temporal and semantic relations in video activity localisation}. In \bibinfo{booktitle}{\emph{ICCV}}. \bibinfo{pages}{7199--7208}.
\newblock


\bibitem[Krishna et~al\mbox{.}(2017a)]%
        {krishna2017dense}
\bibfield{author}{\bibinfo{person}{Ranjay Krishna}, \bibinfo{person}{Kenji Hata}, \bibinfo{person}{Frederic Ren}, \bibinfo{person}{Li Fei-Fei}, {and} \bibinfo{person}{Juan Carlos~Niebles}.} \bibinfo{year}{2017}\natexlab{a}.
\newblock \showarticletitle{Dense-captioning events in videos}. In \bibinfo{booktitle}{\emph{ICCV}}. \bibinfo{pages}{706--715}.
\newblock


\bibitem[Krishna et~al\mbox{.}(2017b)]%
        {krishna2017visual}
\bibfield{author}{\bibinfo{person}{Ranjay Krishna}, \bibinfo{person}{Yuke Zhu}, \bibinfo{person}{Oliver Groth}, \bibinfo{person}{Justin Johnson}, \bibinfo{person}{Kenji Hata}, \bibinfo{person}{Joshua Kravitz}, \bibinfo{person}{Stephanie Chen}, \bibinfo{person}{Yannis Kalantidis}, \bibinfo{person}{Li-Jia Li}, \bibinfo{person}{David~A Shamma}, {et~al\mbox{.}}} \bibinfo{year}{2017}\natexlab{b}.
\newblock \showarticletitle{Visual genome: Connecting language and vision using crowdsourced dense image annotations}.
\newblock \bibinfo{journal}{\emph{International journal of computer vision}}  \bibinfo{volume}{123} (\bibinfo{year}{2017}), \bibinfo{pages}{32--73}.
\newblock


\bibitem[Lei et~al\mbox{.}(2021b)]%
        {lei2021understanding}
\bibfield{author}{\bibinfo{person}{Chenyi Lei}, \bibinfo{person}{Shixian Luo}, \bibinfo{person}{Yong Liu}, \bibinfo{person}{Wanggui He}, \bibinfo{person}{Jiamang Wang}, \bibinfo{person}{Guoxin Wang}, \bibinfo{person}{Haihong Tang}, \bibinfo{person}{Chunyan Miao}, {and} \bibinfo{person}{Houqiang Li}.} \bibinfo{year}{2021}\natexlab{b}.
\newblock \showarticletitle{Understanding chinese video and language via contrastive multimodal pre-training}. In \bibinfo{booktitle}{\emph{ACMMM}}. \bibinfo{pages}{2567--2576}.
\newblock


\bibitem[Lei et~al\mbox{.}(2021a)]%
        {lei2021less}
\bibfield{author}{\bibinfo{person}{Jie Lei}, \bibinfo{person}{Linjie Li}, \bibinfo{person}{Luowei Zhou}, \bibinfo{person}{Zhe Gan}, \bibinfo{person}{Tamara~L Berg}, \bibinfo{person}{Mohit Bansal}, {and} \bibinfo{person}{Jingjing Liu}.} \bibinfo{year}{2021}\natexlab{a}.
\newblock \showarticletitle{Less is more: Clipbert for video-and-language learning via sparse sampling}. In \bibinfo{booktitle}{\emph{CVPR}}. \bibinfo{pages}{7331--7341}.
\newblock


\bibitem[Li et~al\mbox{.}(2022a)]%
        {li2022blip}
\bibfield{author}{\bibinfo{person}{Junnan Li}, \bibinfo{person}{Dongxu Li}, \bibinfo{person}{Caiming Xiong}, {and} \bibinfo{person}{Steven Hoi}.} \bibinfo{year}{2022}\natexlab{a}.
\newblock \showarticletitle{Blip: Bootstrapping language-image pre-training for unified vision-language understanding and generation}.
\newblock \bibinfo{journal}{\emph{arXiv preprint arXiv:2201.12086}} (\bibinfo{year}{2022}).
\newblock


\bibitem[Li et~al\mbox{.}(2022b)]%
        {li2022compositional}
\bibfield{author}{\bibinfo{person}{Juncheng Li}, \bibinfo{person}{Junlin Xie}, \bibinfo{person}{Long Qian}, \bibinfo{person}{Linchao Zhu}, \bibinfo{person}{Siliang Tang}, \bibinfo{person}{Fei Wu}, \bibinfo{person}{Yi Yang}, \bibinfo{person}{Yueting Zhuang}, {and} \bibinfo{person}{Xin~Eric Wang}.} \bibinfo{year}{2022}\natexlab{b}.
\newblock \showarticletitle{Compositional temporal grounding with structured variational cross-graph correspondence learning}. In \bibinfo{booktitle}{\emph{CVPR}}. \bibinfo{pages}{3032--3041}.
\newblock


\bibitem[Li et~al\mbox{.}(2020)]%
        {li2020hero}
\bibfield{author}{\bibinfo{person}{Linjie Li}, \bibinfo{person}{YenChun Chen}, \bibinfo{person}{Yu Cheng}, \bibinfo{person}{Zhe Gan}, \bibinfo{person}{Licheng Yu}, {and} \bibinfo{person}{Jingjing Liu}.} \bibinfo{year}{2020}\natexlab{}.
\newblock \showarticletitle{Hero: Hierarchical encoder for video+ language omni-representation pre-training}.
\newblock \bibinfo{journal}{\emph{arXiv preprint arXiv:2005.00200}} (\bibinfo{year}{2020}).
\newblock


\bibitem[Li et~al\mbox{.}(2022c)]%
        {li2022grounded}
\bibfield{author}{\bibinfo{person}{Liunian~Harold Li}, \bibinfo{person}{Pengchuan Zhang}, \bibinfo{person}{Haotian Zhang}, \bibinfo{person}{Jianwei Yang}, \bibinfo{person}{Chunyuan Li}, \bibinfo{person}{Yiwu Zhong}, \bibinfo{person}{Lijuan Wang}, \bibinfo{person}{Lu Yuan}, \bibinfo{person}{Lei Zhang}, \bibinfo{person}{JenqNeng Hwang}, {et~al\mbox{.}}} \bibinfo{year}{2022}\natexlab{c}.
\newblock \showarticletitle{Grounded language-image pre-training}. In \bibinfo{booktitle}{\emph{CVPR}}. \bibinfo{pages}{10965--10975}.
\newblock


\bibitem[Lin et~al\mbox{.}(2023)]%
        {lin2023univtg}
\bibfield{author}{\bibinfo{person}{Kevin~Qinghong Lin}, \bibinfo{person}{Pengchuan Zhang}, \bibinfo{person}{Joya Chen}, \bibinfo{person}{Shraman Pramanick}, \bibinfo{person}{Difei Gao}, \bibinfo{person}{Alex~Jinpeng Wang}, \bibinfo{person}{Rui Yan}, {and} \bibinfo{person}{Mike~Zheng Shou}.} \bibinfo{year}{2023}\natexlab{}.
\newblock \showarticletitle{Univtg: Towards unified video-language temporal grounding}. In \bibinfo{booktitle}{\emph{Proceedings of the IEEE/CVF International Conference on Computer Vision}}. \bibinfo{pages}{2794--2804}.
\newblock


\bibitem[Lin et~al\mbox{.}(2014)]%
        {lin2014microsoft}
\bibfield{author}{\bibinfo{person}{Tsung-Yi Lin}, \bibinfo{person}{Michael Maire}, \bibinfo{person}{Serge Belongie}, \bibinfo{person}{James Hays}, \bibinfo{person}{Pietro Perona}, \bibinfo{person}{Deva Ramanan}, \bibinfo{person}{Piotr Doll{\'a}r}, {and} \bibinfo{person}{C~Lawrence Zitnick}.} \bibinfo{year}{2014}\natexlab{}.
\newblock \showarticletitle{Microsoft coco: Common objects in context}. In \bibinfo{booktitle}{\emph{Computer Vision--ECCV 2014: 13th European Conference, Zurich, Switzerland, September 6-12, 2014, Proceedings, Part V 13}}. Springer, \bibinfo{pages}{740--755}.
\newblock


\bibitem[Lin et~al\mbox{.}(2020)]%
        {lin2020weakly}
\bibfield{author}{\bibinfo{person}{Zhijie Lin}, \bibinfo{person}{Zhou Zhao}, \bibinfo{person}{Zhu Zhang}, \bibinfo{person}{Qi Wang}, {and} \bibinfo{person}{Huasheng Liu}.} \bibinfo{year}{2020}\natexlab{}.
\newblock \showarticletitle{Weakly-supervised video moment retrieval via semantic completion network}.
\newblock \bibinfo{journal}{\emph{AAAI}} \bibinfo{volume}{34}, \bibinfo{number}{07} (\bibinfo{date}{April} \bibinfo{year}{2020}), \bibinfo{pages}{11539--11546}.
\newblock


\bibitem[Liu et~al\mbox{.}(2022)]%
        {liu2022unsupervised}
\bibfield{author}{\bibinfo{person}{Daizong Liu}, \bibinfo{person}{Xiaoye Qu}, \bibinfo{person}{Yinzhen Wang}, \bibinfo{person}{Xing Di}, \bibinfo{person}{Kai Zou}, \bibinfo{person}{Yu Cheng}, \bibinfo{person}{Zichuan Xu}, {and} \bibinfo{person}{Pan Zhou}.} \bibinfo{year}{2022}\natexlab{}.
\newblock \showarticletitle{Unsupervised temporal video grounding with deep semantic clustering}.
\newblock \bibinfo{journal}{\emph{AAAI}} \bibinfo{volume}{36}, \bibinfo{number}{2} (\bibinfo{date}{June} \bibinfo{year}{2022}), \bibinfo{pages}{1683--1691}.
\newblock


\bibitem[Ma et~al\mbox{.}(2020)]%
        {ma2020vlanet}
\bibfield{author}{\bibinfo{person}{Minuk Ma}, \bibinfo{person}{Sunjae Yoon}, \bibinfo{person}{Junyeong Kim}, \bibinfo{person}{Youngjoon Lee}, \bibinfo{person}{Sunghun Kang}, {and} \bibinfo{person}{Chang~D Yoo}.} \bibinfo{year}{2020}\natexlab{}.
\newblock \showarticletitle{Vlanet: Video-language alignment network for weakly-supervised video moment retrieval}. In \bibinfo{booktitle}{\emph{ECCV}}. Springer, \bibinfo{pages}{156--171}.
\newblock


\bibitem[Miech et~al\mbox{.}(2019)]%
        {miech2019howto100m}
\bibfield{author}{\bibinfo{person}{Antoine Miech}, \bibinfo{person}{Dimitri Zhukov}, \bibinfo{person}{Jean-Baptiste Alayrac}, \bibinfo{person}{Makarand Tapaswi}, \bibinfo{person}{Ivan Laptev}, {and} \bibinfo{person}{Josef Sivic}.} \bibinfo{year}{2019}\natexlab{}.
\newblock \showarticletitle{Howto100m: Learning a text-video embedding by watching hundred million narrated video clips}. In \bibinfo{booktitle}{\emph{ICCV}}. \bibinfo{pages}{2630--2640}.
\newblock


\bibitem[Mithun et~al\mbox{.}(2019)]%
        {mithun2019weakly}
\bibfield{author}{\bibinfo{person}{Niluthpol~Chowdhury Mithun}, \bibinfo{person}{Sujoy Paul}, {and} \bibinfo{person}{Amit~K Roy-Chowdhury}.} \bibinfo{year}{2019}\natexlab{}.
\newblock \showarticletitle{Weakly supervised video moment retrieval from text queries}. In \bibinfo{booktitle}{\emph{CVPR}}. \bibinfo{pages}{11592--11601}.
\newblock


\bibitem[Mun et~al\mbox{.}(2020)]%
        {mun2020local}
\bibfield{author}{\bibinfo{person}{Jonghwan Mun}, \bibinfo{person}{Minsu Cho}, {and} \bibinfo{person}{Bohyung Han}.} \bibinfo{year}{2020}\natexlab{}.
\newblock \showarticletitle{Local-global video-text interactions for temporal grounding}. In \bibinfo{booktitle}{\emph{CVPR}}. \bibinfo{pages}{10810--10819}.
\newblock


\bibitem[Nam et~al\mbox{.}(2021)]%
        {nam2021zero}
\bibfield{author}{\bibinfo{person}{Jinwoo Nam}, \bibinfo{person}{Daechul Ahn}, \bibinfo{person}{Dongyeop Kang}, \bibinfo{person}{Seong~Jong Ha}, {and} \bibinfo{person}{Jonghyun Choi}.} \bibinfo{year}{2021}\natexlab{}.
\newblock \showarticletitle{Zero-shot natural language video localization}. In \bibinfo{booktitle}{\emph{ICCV}}. \bibinfo{pages}{1470--1479}.
\newblock


\bibitem[Radford et~al\mbox{.}(2021)]%
        {radford2021learning}
\bibfield{author}{\bibinfo{person}{Alec Radford}, \bibinfo{person}{Jong~Wook Kim}, \bibinfo{person}{Chris Hallacy}, \bibinfo{person}{Aditya Ramesh}, \bibinfo{person}{Gabriel Goh}, \bibinfo{person}{Sandhini Agarwal}, \bibinfo{person}{Girish Sastry}, \bibinfo{person}{Amanda Askell}, \bibinfo{person}{Pamela Mishkin}, \bibinfo{person}{Jack Clark}, {et~al\mbox{.}}} \bibinfo{year}{2021}\natexlab{}.
\newblock \showarticletitle{Learning transferable visual models from natural language supervision}. In \bibinfo{booktitle}{\emph{ICML}}. PMLR, \bibinfo{pages}{8748--8763}.
\newblock


\bibitem[Rao et~al\mbox{.}(2022)]%
        {rao2022denseclip}
\bibfield{author}{\bibinfo{person}{Yongming Rao}, \bibinfo{person}{Wenliang Zhao}, \bibinfo{person}{Guangyi Chen}, \bibinfo{person}{Yansong Tang}, \bibinfo{person}{Zheng Zhu}, \bibinfo{person}{Guan Huang}, \bibinfo{person}{Jie Zhou}, {and} \bibinfo{person}{Jiwen Lu}.} \bibinfo{year}{2022}\natexlab{}.
\newblock \showarticletitle{Denseclip: Language-guided dense prediction with context-aware prompting}. In \bibinfo{booktitle}{\emph{CVPR}}. \bibinfo{pages}{18082--18091}.
\newblock


\bibitem[Ren et~al\mbox{.}(2015)]%
        {ren2015faster}
\bibfield{author}{\bibinfo{person}{Shaoqing Ren}, \bibinfo{person}{Kaiming He}, \bibinfo{person}{Ross Girshick}, {and} \bibinfo{person}{Jian Sun}.} \bibinfo{year}{2015}\natexlab{}.
\newblock \showarticletitle{Faster r-cnn: Towards real-time object detection with region proposal networks}.
\newblock \bibinfo{journal}{\emph{Advances in neural information processing systems}}  \bibinfo{volume}{28} (\bibinfo{year}{2015}).
\newblock


\bibitem[Sharma et~al\mbox{.}(2018)]%
        {sharma2018conceptual}
\bibfield{author}{\bibinfo{person}{Piyush Sharma}, \bibinfo{person}{Nan Ding}, \bibinfo{person}{Sebastian Goodman}, {and} \bibinfo{person}{Radu Soricut}.} \bibinfo{year}{2018}\natexlab{}.
\newblock \showarticletitle{Conceptual Captions: A Cleaned, Hypernymed, Image Alt-text Dataset For Automatic Image Captioning}. In \bibinfo{booktitle}{\emph{Proceedings of ACL}}.
\newblock


\bibitem[Song et~al\mbox{.}(2020)]%
        {song2020weakly}
\bibfield{author}{\bibinfo{person}{Yijun Song}, \bibinfo{person}{Jingwen Wang}, \bibinfo{person}{Lin Ma}, \bibinfo{person}{Zhou Yu}, {and} \bibinfo{person}{Jun Yu}.} \bibinfo{year}{2020}\natexlab{}.
\newblock \showarticletitle{Weakly-supervised multi-level attentional reconstruction network for grounding textual queries in videos}.
\newblock \bibinfo{journal}{\emph{arXiv preprint arXiv:2003.07048}} (\bibinfo{year}{2020}).
\newblock


\bibitem[Sun et~al\mbox{.}(2019)]%
        {sun2019videobert}
\bibfield{author}{\bibinfo{person}{Chen Sun}, \bibinfo{person}{Austin Myers}, \bibinfo{person}{Carl Vondrick}, \bibinfo{person}{Kevin Murphy}, {and} \bibinfo{person}{Cordelia Schmid}.} \bibinfo{year}{2019}\natexlab{}.
\newblock \showarticletitle{Videobert: A joint model for video and language representation learning}. In \bibinfo{booktitle}{\emph{ICCV}}. \bibinfo{pages}{7464--7473}.
\newblock


\bibitem[Tan et~al\mbox{.}(2021)]%
        {tan2021logan}
\bibfield{author}{\bibinfo{person}{Reuben Tan}, \bibinfo{person}{Huijuan Xu}, \bibinfo{person}{Kate Saenko}, {and} \bibinfo{person}{Bryan~A Plummer}.} \bibinfo{year}{2021}\natexlab{}.
\newblock \showarticletitle{Logan: Latent graph co-attention network for weakly-supervised video moment retrieval}. In \bibinfo{booktitle}{\emph{WACV}}. \bibinfo{pages}{2083--2092}.
\newblock


\bibitem[Tang et~al\mbox{.}(2021)]%
        {tang2021decembert}
\bibfield{author}{\bibinfo{person}{Zineng Tang}, \bibinfo{person}{Jie Lei}, {and} \bibinfo{person}{Mohit Bansal}.} \bibinfo{year}{2021}\natexlab{}.
\newblock \showarticletitle{Decembert: Learning from noisy instructional videos via dense captions and entropy minimization}. In \bibinfo{booktitle}{\emph{NAACL}}. \bibinfo{pages}{2415--2426}.
\newblock


\bibitem[Wang et~al\mbox{.}(2022)]%
        {wang2022prompt}
\bibfield{author}{\bibinfo{person}{Guolong Wang}, \bibinfo{person}{Xun Wu}, \bibinfo{person}{Zhaoyuan Liu}, {and} \bibinfo{person}{Junchi Yan}.} \bibinfo{year}{2022}\natexlab{}.
\newblock \showarticletitle{Prompt-based zero-shot video moment retrieval}. In \bibinfo{booktitle}{\emph{ACMMM}}. \bibinfo{pages}{413--421}.
\newblock


\bibitem[Wang et~al\mbox{.}(2021)]%
        {wang2021structured}
\bibfield{author}{\bibinfo{person}{Hao Wang}, \bibinfo{person}{Zheng-Jun Zha}, \bibinfo{person}{Liang Li}, \bibinfo{person}{Dong Liu}, {and} \bibinfo{person}{Jiebo Luo}.} \bibinfo{year}{2021}\natexlab{}.
\newblock \showarticletitle{Structured multi-level interaction network for video moment localization via language query}. In \bibinfo{booktitle}{\emph{CVPR}}. \bibinfo{pages}{7026--7035}.
\newblock


\bibitem[Wang et~al\mbox{.}(2023)]%
        {wang2023mixup}
\bibfield{author}{\bibinfo{person}{Xin Wang}, \bibinfo{person}{Zihao Wu}, \bibinfo{person}{Hong Chen}, \bibinfo{person}{Xiaohan Lan}, {and} \bibinfo{person}{Wenwu Zhu}.} \bibinfo{year}{2023}\natexlab{}.
\newblock \showarticletitle{Mixup-Augmented Temporally Debiased Video Grounding with Content-Location Disentanglement}. In \bibinfo{booktitle}{\emph{Proceedings of the 31st ACM International Conference on Multimedia}}. \bibinfo{pages}{4450--4459}.
\newblock


\bibitem[Xiao et~al\mbox{.}(2021)]%
        {xiao2021boundary}
\bibfield{author}{\bibinfo{person}{Shaoning Xiao}, \bibinfo{person}{Long Chen}, \bibinfo{person}{Songyang Zhang}, \bibinfo{person}{Wei Ji}, \bibinfo{person}{Jian Shao}, \bibinfo{person}{Lu Ye}, {and} \bibinfo{person}{Jun Xiao}.} \bibinfo{year}{2021}\natexlab{}.
\newblock \showarticletitle{Boundary proposal network for two-stage natural language video localization}.
\newblock \bibinfo{journal}{\emph{AAAI}} \bibinfo{volume}{35}, \bibinfo{number}{4} (\bibinfo{date}{May} \bibinfo{year}{2021}), \bibinfo{pages}{2986--2994}.
\newblock


\bibitem[Xu et~al\mbox{.}(2021a)]%
        {xu2021vlm}
\bibfield{author}{\bibinfo{person}{Hu Xu}, \bibinfo{person}{Gargi Ghosh}, \bibinfo{person}{Po-Yao Huang}, \bibinfo{person}{Prahal Arora}, \bibinfo{person}{Masoumeh Aminzadeh}, \bibinfo{person}{Christoph Feichtenhofer}, \bibinfo{person}{Florian Metze}, {and} \bibinfo{person}{Luke Zettlemoyer}.} \bibinfo{year}{2021}\natexlab{a}.
\newblock \showarticletitle{VLM: Task-agnostic video-language model pre-training for video understanding}.
\newblock \bibinfo{journal}{\emph{arXiv preprint arXiv:2105.09996}} (\bibinfo{year}{2021}).
\newblock


\bibitem[Xu et~al\mbox{.}(2019)]%
        {xu2019multilevel}
\bibfield{author}{\bibinfo{person}{Huijuan Xu}, \bibinfo{person}{Kun He}, \bibinfo{person}{Bryan~A. Plummer}, \bibinfo{person}{Leonid Sigal}, \bibinfo{person}{Stan Sclaroff}, {and} \bibinfo{person}{Kate Saenko}.} \bibinfo{year}{2019}\natexlab{}.
\newblock \showarticletitle{Multilevel language and vision integration for text-to-clip retrieval}.
\newblock \bibinfo{journal}{\emph{AAAI}} \bibinfo{volume}{33}, \bibinfo{number}{01} (\bibinfo{date}{July} \bibinfo{year}{2019}), \bibinfo{pages}{9062--9069}.
\newblock


\bibitem[Xu et~al\mbox{.}(2021b)]%
        {xu2021boundary}
\bibfield{author}{\bibinfo{person}{Mengmeng Xu}, \bibinfo{person}{Juan-Manuel P{\'e}rez-R{\'u}a}, \bibinfo{person}{Victor Escorcia}, \bibinfo{person}{Brais Martinez}, \bibinfo{person}{Xiatian Zhu}, \bibinfo{person}{Li Zhang}, \bibinfo{person}{Bernard Ghanem}, {and} \bibinfo{person}{Tao Xiang}.} \bibinfo{year}{2021}\natexlab{b}.
\newblock \showarticletitle{Boundary-sensitive pre-training for temporal localization in videos}. In \bibinfo{booktitle}{\emph{ICCV}}. \bibinfo{pages}{7220--7230}.
\newblock


\bibitem[Xu et~al\mbox{.}(2021c)]%
        {xu2021low}
\bibfield{author}{\bibinfo{person}{Mengmeng Xu}, \bibinfo{person}{Juan~Manuel Perez~Rua}, \bibinfo{person}{Xiatian Zhu}, \bibinfo{person}{Bernard Ghanem}, {and} \bibinfo{person}{Brais Martinez}.} \bibinfo{year}{2021}\natexlab{c}.
\newblock \showarticletitle{Low-fidelity video encoder optimization for temporal action localization}.
\newblock \bibinfo{journal}{\emph{Advances in NIPS}}  \bibinfo{volume}{34} (\bibinfo{year}{2021}), \bibinfo{pages}{9923--9935}.
\newblock


\bibitem[Yang et~al\mbox{.}(2021)]%
        {yang2021local}
\bibfield{author}{\bibinfo{person}{Wenfei Yang}, \bibinfo{person}{Tianzhu Zhang}, \bibinfo{person}{Yongdong Zhang}, {and} \bibinfo{person}{Feng Wu}.} \bibinfo{year}{2021}\natexlab{}.
\newblock \showarticletitle{Local correspondence network for weakly supervised temporal sentence grounding}.
\newblock \bibinfo{journal}{\emph{IEEE TIP}}  \bibinfo{volume}{30} (\bibinfo{year}{2021}), \bibinfo{pages}{3252--3262}.
\newblock


\bibitem[Yuan et~al\mbox{.}(2021)]%
        {yuan2021closer}
\bibfield{author}{\bibinfo{person}{Yitian Yuan}, \bibinfo{person}{Xiaohan Lan}, \bibinfo{person}{Xin Wang}, \bibinfo{person}{Long Chen}, \bibinfo{person}{Zhi Wang}, {and} \bibinfo{person}{Wenwu Zhu}.} \bibinfo{year}{2021}\natexlab{}.
\newblock \showarticletitle{A closer look at temporal sentence grounding in videos: Dataset and metric}. In \bibinfo{booktitle}{\emph{Proceedings of the 2nd International Workshop on Human-centric Multimedia Analysis}}. \bibinfo{pages}{13--21}.
\newblock


\bibitem[Yuan et~al\mbox{.}(2019)]%
        {yuan2019find}
\bibfield{author}{\bibinfo{person}{Yitian Yuan}, \bibinfo{person}{Tao Mei}, {and} \bibinfo{person}{Wenwu Zhu}.} \bibinfo{year}{2019}\natexlab{}.
\newblock \showarticletitle{To find where you talk: temporal sentence localization in video with attention based location regression}.
\newblock \bibinfo{journal}{\emph{AAAI}} \bibinfo{volume}{33}, \bibinfo{number}{01} (\bibinfo{date}{July} \bibinfo{year}{2019}), \bibinfo{pages}{9159--9166}.
\newblock


\bibitem[Zhang et~al\mbox{.}(2022)]%
        {zhang2022unsupervised}
\bibfield{author}{\bibinfo{person}{Can Zhang}, \bibinfo{person}{Tianyu Yang}, \bibinfo{person}{Junwu Weng}, \bibinfo{person}{Meng Cao}, \bibinfo{person}{Jue Wang}, {and} \bibinfo{person}{Yuexian Zou}.} \bibinfo{year}{2022}\natexlab{}.
\newblock \showarticletitle{Unsupervised pre-training for temporal action localization tasks}. In \bibinfo{booktitle}{\emph{CVPR}}. \bibinfo{pages}{14031--14041}.
\newblock


\bibitem[Zhang et~al\mbox{.}(2019)]%
        {zhang2019man}
\bibfield{author}{\bibinfo{person}{Da Zhang}, \bibinfo{person}{Xiyang Dai}, \bibinfo{person}{Xin Wang}, \bibinfo{person}{YuanFang Wang}, {and} \bibinfo{person}{Larry~S Davis}.} \bibinfo{year}{2019}\natexlab{}.
\newblock \showarticletitle{Man: Moment alignment network for natural language moment retrieval via iterative graph adjustment}. In \bibinfo{booktitle}{\emph{CVPR}}. \bibinfo{pages}{1247--1257}.
\newblock


\bibitem[Zhang et~al\mbox{.}(2020a)]%
        {zhang2020learning}
\bibfield{author}{\bibinfo{person}{Songyang Zhang}, \bibinfo{person}{Houwen Peng}, \bibinfo{person}{Jianlong Fu}, {and} \bibinfo{person}{Jiebo Luo}.} \bibinfo{year}{2020}\natexlab{a}.
\newblock \showarticletitle{Learning 2D temporal adjacent networks for moment localization with natural language}.
\newblock \bibinfo{journal}{\emph{AAAI}} \bibinfo{volume}{34}, \bibinfo{number}{07} (\bibinfo{date}{April} \bibinfo{year}{2020}), \bibinfo{pages}{12870--12877}.
\newblock


\bibitem[Zhang et~al\mbox{.}(2020b)]%
        {zhang2020counterfactual}
\bibfield{author}{\bibinfo{person}{Zhu Zhang}, \bibinfo{person}{Zhou Zhao}, \bibinfo{person}{Zhijie Lin}, \bibinfo{person}{Xiuqiang He}, {et~al\mbox{.}}} \bibinfo{year}{2020}\natexlab{b}.
\newblock \showarticletitle{Counterfactual contrastive learning for weakly-supervised vision-language grounding}.
\newblock \bibinfo{journal}{\emph{Advances in NIPS}}  \bibinfo{volume}{33} (\bibinfo{year}{2020}), \bibinfo{pages}{18123--18134}.
\newblock


\bibitem[Zhou et~al\mbox{.}(2022a)]%
        {zhou2022conditional}
\bibfield{author}{\bibinfo{person}{Kaiyang Zhou}, \bibinfo{person}{Jingkang Yang}, \bibinfo{person}{Chen~Change Loy}, {and} \bibinfo{person}{Ziwei Liu}.} \bibinfo{year}{2022}\natexlab{a}.
\newblock \showarticletitle{Conditional prompt learning for vision-language models}. In \bibinfo{booktitle}{\emph{CVPR}}. \bibinfo{pages}{16816--16825}.
\newblock


\bibitem[Zhou et~al\mbox{.}(2022b)]%
        {zhou2022learning}
\bibfield{author}{\bibinfo{person}{Kaiyang Zhou}, \bibinfo{person}{Jingkang Yang}, \bibinfo{person}{Chen~Change Loy}, {and} \bibinfo{person}{Ziwei Liu}.} \bibinfo{year}{2022}\natexlab{b}.
\newblock \showarticletitle{Learning to prompt for vision-language models}.
\newblock \bibinfo{journal}{\emph{IJCV}} \bibinfo{volume}{130}, \bibinfo{number}{9} (\bibinfo{year}{2022}), \bibinfo{pages}{2337--2348}.
\newblock


\bibitem[Zhu and Yang(2020)]%
        {zhu2020actbert}
\bibfield{author}{\bibinfo{person}{Linchao Zhu} {and} \bibinfo{person}{Yi Yang}.} \bibinfo{year}{2020}\natexlab{}.
\newblock \showarticletitle{Actbert: Learning global-local video-text representations}. In \bibinfo{booktitle}{\emph{CVPR}}. \bibinfo{pages}{8746--8755}.
\newblock


\end{thebibliography}
